%% file: main.tex
\documentclass[acmsmall,screen=true,anonymous=false,bookmarks=false]{acmart}
\input{set_acm}

\usepackage{times}
\usepackage{listings}
\usepackage{tabularx}
\usepackage[linesnumbered,ruled,vlined]{algorithm2e}

\setcopyright{rightsretained}
\acmJournal{TODAES}
\acmYear{2026} \acmVolume{1} \acmNumber{1} \acmArticle{}
\acmMonth{1} \acmDOI{10.1145/3801559}

\begin{document}

% \onecolumn
% \input{section/cover}   \clearpage
% \setcounter{page}{1}

\title[Hardware-aware Low-Rank Adaptation for Large Language Models Based on Hybrid CIM]{Hardware-aware Low-Rank Adaptation for Large Language Models Based on Hybrid Compute-in-Memory Architecture}

\author{Taiqiang Wu}
\authornote{These authors contributed equally.}
\affiliation{\institution{The University of Hong Kong}}
\author{Chenchen Ding}
\authornotemark[1]
\affiliation{\institution{The University of Hong Kong}}
\author{Wenyong Zhou}
\authornotemark[1]
\affiliation{\institution{The University of Hong Kong}}
\author{Yuxin Cheng}   \affiliation{\institution{The University of Hong Kong}}
\author{Xincheng Feng}   \affiliation{\institution{The University of Hong Kong}}
\author{Shuqi Wang}   \affiliation{\institution{The University of Hong Kong}}
\author{Wendong Xu}   \affiliation{\institution{The University of Hong Kong}}
\author{Chufan Shi}   \affiliation{\institution{Tsinghua University}}
\author{Zhengwu Liu}   
\authornote{Corresponding author}
\affiliation{\institution{The University of Hong Kong}}
% \authornotemark[2]
\author{Ngai Wong}   
\authornotemark[2]
\affiliation{\institution{The University of Hong Kong}}

% \shortauthors{Taiqiang Wu, Chenchen Ding, Wenyong Zhou}

% \author{Taiqiang Wu$^\dag$, Chenchen Ding$^\dag$, Wenyong Zhou$^\dag$, \\ Yuxin Cheng, Xincheng Feng, Shuqi Wang, Chufan Shi, Zhengwu Liu*, and Ngai Wong*
% \thanks{
% This work was supported in part by the Theme-based Research Scheme (TRS) project T45-701/22-R, National Natural Science Foundation of China (62404187), and the General Research Fund (GRF) Project 17203224, of the Research Grants Council (RGC), Hong Kong SAR.

% Chufan is from Tsinghua University, and the other authors are with the Department of Electrical and Electronic Engineering, The University of Hong Kong.
% $^\dag$ for equal contributions.
% *Corresponding author: Zhengwu Liu, Ngai Wong, E-mail:\{zwliu, nwong\}@eee.hku.hk.
% }
% }
\renewcommand{\shortauthors}{Taiqiang Wu, Chenchen Ding, Wenyong Zhou, Zhengwu Liu, Ngai Wong, et al.}

\newcommand{\re}[1]{\textcolor{black}{#1}}

\begin{abstract}
Low-rank adaptation (LoRA) is a predominant parameter-efficient finetuning method for adapting large language models (LLMs) to downstream tasks. 
Meanwhile, Compute-in-Memory (CIM) architectures demonstrate superior energy efficiency due to their array-level parallel in-memory computing designs.
In this paper, we propose deploying the LoRA-finetuned LLMs on the hybrid CIM architecture (i.e., pretrained weights onto energy-efficient Resistive Random-Access Memory (RRAM) and LoRA branches onto noise-free Static Random-Access Memory (SRAM)), reducing the energy cost to about 3\% compared to the Nvidia A100 GPU.
However, the inherent noise of RRAM on the saved weights leads to performance degradation, simultaneously.
To address this issue, we design a novel Hardware-aware Low-rank Adaptation (HaLoRA) method.
The key insight is to train a LoRA branch that is robust toward such noise and then deploy it on noise-free SRAM, while the extra cost is negligible since the parameters of LoRAs are much fewer than pretrained weights (e.g., 0.15\% for LLaMA-3.2 1B model).
To improve the robustness towards the noise, we theoretically analyze the gap between the optimization trajectories of the LoRA branch under both ideal and noisy conditions and further design an extra loss to minimize the upper bound of this gap.
Therefore, we can enjoy both energy efficiency and accuracy during inference.
Experiments finetuning the Qwen and LLaMA series demonstrate the effectiveness of HaLoRA across multiple reasoning tasks, achieving up to \textbf{22.7} improvement in average score while maintaining robustness at various noise types and noise levels.
\end{abstract}

\maketitle

\input{section/1-Introduction}

\input{section/2-Related_work}
\input{section/3-Methodology}

\input{section/4-Experiments}
\input{section/5-Conclusion}

% \bibliographystyle{IEEEtran}
% \bibliography{reference}
\input{main.bbl}

% \clearpage
% \input{section/answer}

\end{document}

%% file: set_acm.tex
\usepackage{lipsum}
\usepackage{graphicx}
\usepackage{footnote}

\usepackage{amssymb}

\usepackage{footnote}
\usepackage{amsmath}
\usepackage{mathtools}
\usepackage{mathrsfs}                                      % mathscr command
\usepackage{siunitx} 
\usepackage{comment}
\usepackage{wrapfig} 
\usepackage[subrefformat=parens,labelformat=parens]{subfig}
\usepackage{bm,bbm}
\usepackage{multirow}
\usepackage{makecell}
\usepackage{threeparttable,booktabs}
\usepackage{blkarray}
\usepackage{tikz}
\usetikzlibrary{positioning,calc,fit,decorations.pathmorphing,shapes.geometric,shapes.gates.logic.US,calc}
\usepackage{tikz-3dplot}
\usepackage{balance}
\usepackage{courier}                                       % courier font, used in \texttt
\usepackage{pifont}                                        % ding commend
\usepackage{cleveref}                                      % smart citation
\Crefformat{figure}{Fig.~#2#1#3}                           % "Fig.", instead of "Figure"
\Crefname{subfigure}{Fig.}{Figs.}
\Crefname{figure}{Fig.}{Figs.}
\usepackage[mathcal]{eucal}
\usepackage{filecontents}                                  % support to pgfplots
\usepackage{pgfplots}
\usepackage{pgfplotstable}
\usepackage{pgf-pie}
\usepgfplotslibrary{groupplots}
\pgfplotsset{compat=newest}
\usepackage[figuresright]{rotating}
\usepackage{xcolor,colortbl}                               % \rowcolor
\usepackage[inline]{enumitem}                              % inline enumerate
\setlist{leftmargin=5.08mm}

\theoremstyle{plain}

\theoremstyle{definition}

\usepackage[skip=1pt]{caption}            % set space between figure and caption
\definecolor{CUHKorange}{RGB}{244,106,18} %F47012
\definecolor{CUHKblue}{RGB}{0,111,190}    %006FBE
\definecolor{CUHKgreen}{RGB}{0,127,128}   %007F80
\definecolor{CUHKred}{RGB}{228,46,36}     %E42E24
\definecolor{CUHKyellow}{RGB}{198,148,34} %C69422
\definecolor{CUHKdark}{RGB}{114,44,114}   %722C72
\definecolor{CUHKmiddle}{RGB}{144,44,144} %902C90

\usepackage{tcolorbox}
\tcbuselibrary{skins,breakable}
    {\endtcolorbox}

    {\endtcolorbox}

\usepackage{amsfonts}
\usepackage{algorithmic}
\usepackage{array}
\usepackage{textcomp}
\usepackage{stfloats}
\usepackage{url}
\usepackage{verbatim}
\usepackage[table,xcdraw]{xcolor}
\definecolor{gg}{HTML}{e2f0cb}
\usepackage{enumerate}
\definecolor{darkgreen}{RGB}{50,100,0}
\definecolor{darkred}{RGB}{200, 0, 0}
\newcommand{\ccmark}{\textcolor{darkgreen}{\ding{51}}}%
\newcommand{\xxmark}{\textcolor{darkred}{\ding{55}}}%
\definecolor{ForestGreen}{rgb}{0.13, 0.55, 0.13}

%% file: section/1-Introduction.tex
\section{Introduction}

Large language models (LLMs), such as GPT-4 \cite{openai_gpt4}, LLaMA \cite{llama_report}, and Qwen \cite{qwen}, have demonstrated promising performance in various Natural Language Processing (NLP) tasks.
However, this success, primarily driven by the massive number of model parameters, presents two critical challenges for practical applications.
First, adapting LLMs to downstream tasks by updating all parameters requires prohibitive computational resources \cite{wu_moslora,xie2024me}.
Second, model inference demands substantial energy consumption, limiting the widespread deployment of these models \cite{wu2025rethinking, samsi2023words}.

Fortunately, various parameter-efficient finetuning (PEFT) methods have been proposed to finetune the models more efficiently \cite{hou_adapter,Li_prompt,zaken_bitfit}. 
Among these methods, low-rank adaptation (LoRA) \cite{hu_lora} has gained increasing popularity due to its simplicity and efficacy by updating only an extra low-rank matrix while preserving the original pretrained weights. 
Meanwhile, Compute-in-Memory (CIM) architectures perform computations directly within memory arrays, achieving high energy efficiency through array-level parallel computing. 
Deploying models on CIM architecture is a promising way to improve inference efficiency \cite{yao_nature,SRAM_MAC}.
There are two typical memory technologies in CIM, including Resistive Random-Access Memory (RRAM) and Static Random-Access Memory (SRAM).
RRAM-based CIM offers better energy efficiency and computational density but introduces noise due to the non-ideality \cite{liu_survey}. 
In contrast, SRAM-based CIM provides greater reliability but demands a larger silicon footprint and relatively higher power consumption for conventional vector-matrix multiplications \cite{chang_survey}.
% Resistive random-access memory (RRAM) and static random-access memory (SRAM) are two typical memory devices used in CIM implementations, where RRAM is more energy-saving but introduces noise on the saved weights.
Existing work related to CIM has investigated the implementation of small-scale neural networks \cite{RRAM_SRAM_TCAD,chang_science,chang_nature}. 
However, deploying LoRA-finetuned LLMs on CIM architectures remains largely unexplored, presenting challenges and opportunities.

\begin{figure*}[t!]
\centering
\begin{minipage}[h]{0.53\textwidth}
%%%
\centering
\captionof{table}{The comparison between the pretrained weight and LoRA branch when finetuning LLMs such as LLaMA-3.2 1B.}
% \vspace{1em}
\resizebox{\linewidth}{!}
{
\begin{tabular}{l|cc}
\toprule
    & \textbf{Pretrained Weight} & \textbf{LoRA Branch} \\
    \midrule
    Parameters & 1235.8M & 1.9M \\
    Task-agnostic & \ccmark & \xxmark \\
    Trainable & \xxmark & \ccmark \\
    \midrule
    CIM & RRAM & SRAM \\
    Density & High & Low \\
    Accurate & \xxmark & \ccmark \\
\bottomrule
\end{tabular}
}
\label{tab-compra}
\end{minipage}
\quad % 间隔
\begin{minipage}[h]{0.42\textwidth}
%%% 
\centering
\includegraphics[width=\linewidth]{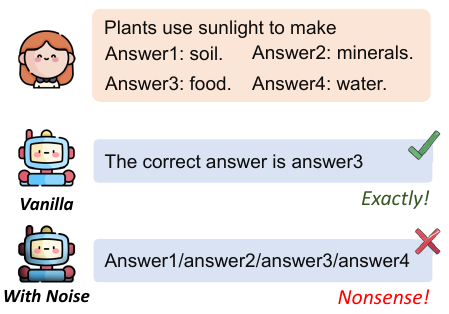}
\caption{One case during inference for ideal and noisy LoRA-finetuned LLMs.
% The non-ideality of RRAM imposes noise on pretrained weights and thus hurts the performance.
}
\label{intro_fig}
\end{minipage}
\end{figure*}

% In this paper, we first propose a hybrid deployment strategy that leverages the complementary advantages of RRAM-based and SRAM-based CIM architectures. 
% Specifically, RRAM-only CIM achieves high energy efficiency but suffers from inherent noise characteristics and complex write-verify operations \cite{ne_cim}, while SRAM-only CIM provides noise-free computation but is limited by its volatile nature and low storage density \cite{nn_cim}.
In this paper, we propose deploying finetuned LLMs on hybrid CIM, leveraging both the energy efficiency and computational density of RRAM and the noise-free computation of SRAM \cite{dhingra2025atleus, sridharan2023x}. 
While the RRAM-only strategy suffers from inherent noise and complex write-verify operations \cite{ne_cim}, the SRAM-only strategy is limited by its volatility and low storage density \cite{nn_cim}.
LoRA-finetuned LLMs exhibit a distinctive structural characteristic: as shown in Table \ref{tab-compra}, the pretrained weights (e.g., 1235.8M parameters in LLaMA-3.2 1B) dominate the model size compared to the LoRA branch (1.9M parameters). 
Our proposed hybrid strategy exploits this characteristic, and thus 1) deploying the task-agnostic pretrained weights on RRAM maximizes energy efficiency while avoiding frequent write operations, and 2) implementing the task-specific LoRA branches on SRAM ensures accurate adaptation at a reasonable cost. 
Despite RRAM's energy efficiency, the inherent non-ideality would introduce noise during the reading process, misleading the model to generate nonsense responses.
As shown in Fig. \ref{intro_fig}, the noise from the RRAM-based weights leads to nonsense answers while the vanilla model can generate the correct answer. 
Hence, these raise an intriguing question: \textit{Can we leverage the accurate LoRA computations on SRAM to compensate for the noise-induced errors from pretrained weights deployed on RRAM, thereby achieving an optimal balance between energy efficiency and task adaptation performance?}

% \input{figures_tables/Table_intro}

% \begin{figure}[!t]
% \centering
% \includegraphics[width=0.6\linewidth]{figures_tables/Figure1_sub.pdf}
% \caption{
% One case during inference for ideal and noisy LoRA-finetuned LLMs. The non-ideality of RRAM imposes noise on pretrained weights and thus hurts the performance.}
% \label{intro_fig}
% % \vspace{-1em}
% \end{figure}

% HaLoRA
To this end, we further propose Hardware-aware Low-Rank Adaptation (HaLoRA), a noise-robust adaptation method. 
% Our key insight is to minimize the discrepancy of two LoRA optimization trajectories with and without noise in pretrained weights during training, aiming to train an optimized LoRA that is both \textbf{robust} and \textbf{accurate} during the inference.
Our key insight is to minimize the discrepancy between LoRA optimization trajectories under ideal and noisy conditions during training, aiming to train an optimized LoRA that is both \textbf{robust} and \textbf{accurate} during the inference.
Specifically, we first inject random noise into the pretrained weights and then optimize the LoRA branches towards the noise-free optimal to improve their robustness.
After that, we theoretically indicate the gap between optimized LoRAs with and without noise and get its \textit{noise-agnostic} upper bound to avoid overfitting to specific noise patterns.
With an extra loss minimizing this upper bound during finetuning, HaLoRA achieves superior robustness and accuracy given noisy RRAM-based pretrained weights during inference.

% 实验&贡献
% We conduct comprehensive experiments by finetuning Qwen 2.5 (0.5B) and LLaMA-3.2 (1B and 3B variants) on commonsense reasoning tasks.
% To evaluate robustness, we inject noise into pretrained weights at three different levels during inference, with each condition tested across five random seeds ranging from 1 to 5. 
We conduct comprehensive experiments by finetuning Qwen2.5 (0.5B) and LLaMA-3.2 (1B and 3B) on commonsense reasoning tasks. 
To evaluate robustness, we inject noise into pretrained weights at three levels, testing each condition with five random seeds.
Experimental results demonstrate that HaLoRA consistently outperforms vanilla LoRA regarding accuracy and robustness.
In particular, at a noise level of 0.02, HaLoRA achieves an average score of 63.1 when finetuning LLaMA-3.2 1B, surpassing LoRA by a significant margin of 22.7 points. 
These results validate our approach's effectiveness in deploying LLMs on hybrid CIM architectures while maintaining robust performance.
Meanwhile, for the LLaMA-3.2 1B with 512 input tokens, HaLoRA achieves 18.1 mJ energy cost, which is 3.29\% of the 550.5 mJ of vanilla LoRA on Nvidia A100 GPU.
Further analyses demonstrate that HaLoRA also performs well considering stuck-at faults.
Our contributions can be summarized as follows:
\begin{itemize}
    \item We propose a novel framework to deploy LoRA-finetuned LLMs on the hybrid CIM architecture, i.e., task-agnostic pretrained weights onto energy-efficient RRAM and task-specific LoRA onto noise-free SRAM.
    \item 
    We introduce HaLoRA to address RRAM non-ideality issues, i.e., the inherited noise on saved weights.
    The key insight is to minimize the gap between the noisy and noise-free training trajectories.
    \item 
    We evaluate HaLoRA by finetuning Qwen2.5 and LLaMA-3.2 on 6 popular commonsense reasoning tasks, demonstrating the effectiveness and robustness at various noise types and noise levels.
\end{itemize}

%% file: section/2-Related_work.tex
\section{Related Work}
\subsection{LoRA and its Variants}
PEFT methods aim to update a small proportion of parameters to adapt LLMs for downstream tasks \cite{hou_adapter,Li_prompt,zaken_bitfit, wu_moslora, hu_lora}.
Among these methods, LoRA \cite{hu_lora}, which injects trainable low-rank branches to approximate the weight updates, has become increasingly popular as it introduces no latency during inference.
In the vanilla LoRA method, the authors introduce two linear projection layers and initialize them as Kaiming uniform and zero matrices \cite{hu_lora}. 
The following variants can be categorized into: 1) searching ranks \cite{zhang_adalora, valipour2022dylora, zhang2024autolora, mao2024dora}; 2) introducing training skills such as setting different learning rates \cite{Hayou_lora+} and improving initialization \cite{meng2024pissa, wang2024milora, wang2024loraga}; and 3) designing new structures.

All these variants focus on improving performance for the ideal scenarios without weight noise.
In this paper, we propose HaLoRA, which is customized for hardware deployment.

\subsection{Hybrid CIM Architecture}
\label{sec:related_cim}
Hybrid CIM architectures combine different memory devices to achieve capabilities beyond what pure single-memory-device architectures can offer. Among them, RRAM-SRAM hybrid architectures have attracted significant attention by combining the high energy efficiency of RRAM with accurate computation of SRAM \cite{wen_science,vlsi_minotaur,liu_hardsea,RRAM_SRAM_TCAD}.
These hybrid designs typically partition computational tasks based on the characteristics of each memory device: deploying high-precision, frequently updated operations on SRAM while allocating computation-intensive yet structurally simple operations to RRAM \cite{wen_science, RRAM_SRAM_TCAD}.
This strategy has facilitated the efficient implementation of convolutional neural networks (CNNs) and lightweight neural architectures \cite{vlsi_minotaur,cnn_cim,chimera}, enabling their deployment in edge computing applications such as robotic localization \cite{slam}, target tracking \cite{tracking}, and recommendation systems \cite{edge_nlp}.

\re{
Existing hybrid architectures primarily focus on implementing small-scale models, with limited exploration of large language models. 
In this work, we explore efficient LLM deployment with a hybrid CIM architecture.
Although at the current stage the massive parameter count of LLMs poses a challenge for on-chip storage, the rapid advancement of RRAM macro capacity (e.g., Mb-level analog macros~\cite{Mb_level}) and system-level scaling techniques, such as 3D stacking~\cite{3D_inte}, multi-core integration~\cite{wan2022compute}, and high-speed inter-chip interconnects~\cite{pcb_level}, provide a clear path toward GB-level system capacity. 
}

\subsection{Robustness Methods against Hardware Non-idealities}

The robustness methods against the RRAM non-idealities have been a hot topic for the past decade.
Specifically, these robust methods can be categorized into 1) noise-aware training, which typically incorporates noise during the training process or introduces robust loss functions \cite{kd_rram, bayes, bayesft}, and 2) hardware compensation strategies, such as mapping critical weights to low-variation areas~\cite{tfix,victor}.
However, these methods mainly focus on the robustness of convolutional neural networks (CNNs) and are \textit{difficult to generalize to LLMs}.
Considering noise-aware training methods, the key is to continuously train the models to improve their robustness, including knowledge distillation \cite{kd_rram} and Bayesian neural network training \cite{bayes, bayesft}.
Due to the massive size of the LLM model, such as 3 billion parameters \cite{llama_report}, the cost of continuous full-parameter training is unaffordable.
Meanwhile, the hardware compensation strategies are impractical for LLMs since pre-testing and correcting each layer through input regularization and column-shared factors introduce substantial additional hardware overhead.

In this paper, we propose HaLoRA to improve the robustness of LoRA-finetuned LLMs at the finetuning stage with negligible extra costs.

%% file: section/3-Methodology.tex
\section{Methodology}

This section first introduces the preliminaries about Transformer and LoRA, followed by the details of the deployment strategy and training process of the proposed HaLoRA method.

\begin{figure*}[!t]
\centering
\includegraphics[width=0.98\linewidth]{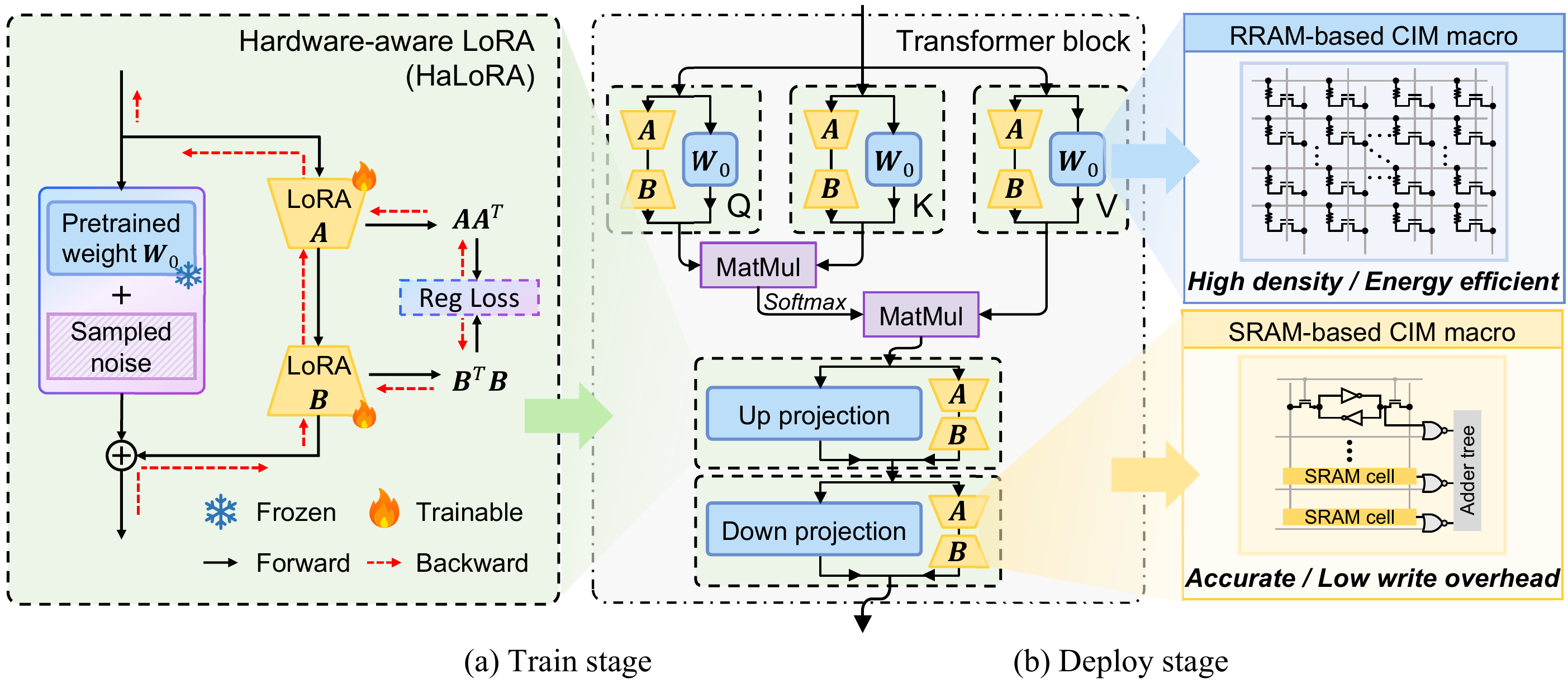}
\caption{
\re{The train and deploy stages for the proposed HaLoRA. (a) During the training stage, HaLoRA incorporates an additional loss regularization term with sampled noise to enhance model robustness. (b) In the deploy stage, the finetuned LLM is mapped to a hybrid CIM architecture formed by RRAM and SRAM-based CIM macros, leveraging their respective advantages.}
}
\label{method_fig}
\end{figure*}

\subsection{Preliminary}
\subsubsection*{Transformer} Transformer \cite{vas_trans} is the predominant architecture for LLMs, such as LLaMA \cite{llama_report} and Qwen \cite{qwen}.
Each transformer layer consists of a multi-head self-attention (MHA) sub-layer and a feed-forward (FFN) sub-layer.
In the MHA sublayer, the input would be projected into K, Q, and V vectors, followed by the nonparametric operations.
Then, the FFN sublayer contains an up-projection linear layer and a down-projection linear layer.
We refer the readers to \cite{wu_wid} for more details.

\subsubsection*{Vanilla LoRA}

Based on the observation that the update in pretrained weights during model adaptation exhibits low intrinsic rank, the LoRA \cite{hu_lora} method aims to model the weight update $\Delta \mathbf{W}$ of weight $\mathbf{W}_0 \in \mathbb{R}^{d_1 \times d_2}$ via two low-rank matrices following:
\begin{equation}
\label{eq_lora}
    \mathbf{W} = \mathbf{W}_0+\Delta \mathbf{W}
\end{equation}
\begin{equation}
\Delta \mathbf{W} = \mathbf{A} \mathbf{B} \in \mathbb{R}^{d_1 \times d_2},
\end{equation}
where $\mathbf{A} \in \mathbb{R}^{d_1 \times r}$ and $\mathbf{B} \in \mathbb{R}^{r \times d_2}$.
The parameters to update would be much fewer since we have $(d_1+d_2)r \ll d_1 \times d_2$ when $r \ll \min(d_1, d_2)$.
During finetuning, we would update the LoRA branch~(i.e., $\mathbf{A}$ and $\mathbf{B}$) while freezing the original pretrained weight $\mathbf{W}_0$.
Therefore, LoRA branches are task-specific while original pretrained weights are task-agnostic.

\subsection{Hybrid CIM Strategy}
\label{section:hybrid_cim}

In this paper, we propose a deployment strategy for finetuned LLMs in a hybrid CIM architecture that leverages the complementary strengths of RRAM and SRAM. 
As shown in Fig.~\ref{method_fig}b, this strategy implements the LLM backbone in RRAM, leveraging its storage density and energy efficiency. 
The noise-sensitive and task-specific LoRA branch is deployed on SRAM to ensure accurate write operations and efficient task adaptation. 
% The extra cost is negligible since the parameters of LoRA are far less than pretrained weights.
% We also provide a detailed comparison in Table \ref{tab-hard}.
\begin{figure*}[!t]
\centering
\includegraphics[width=0.88\linewidth]{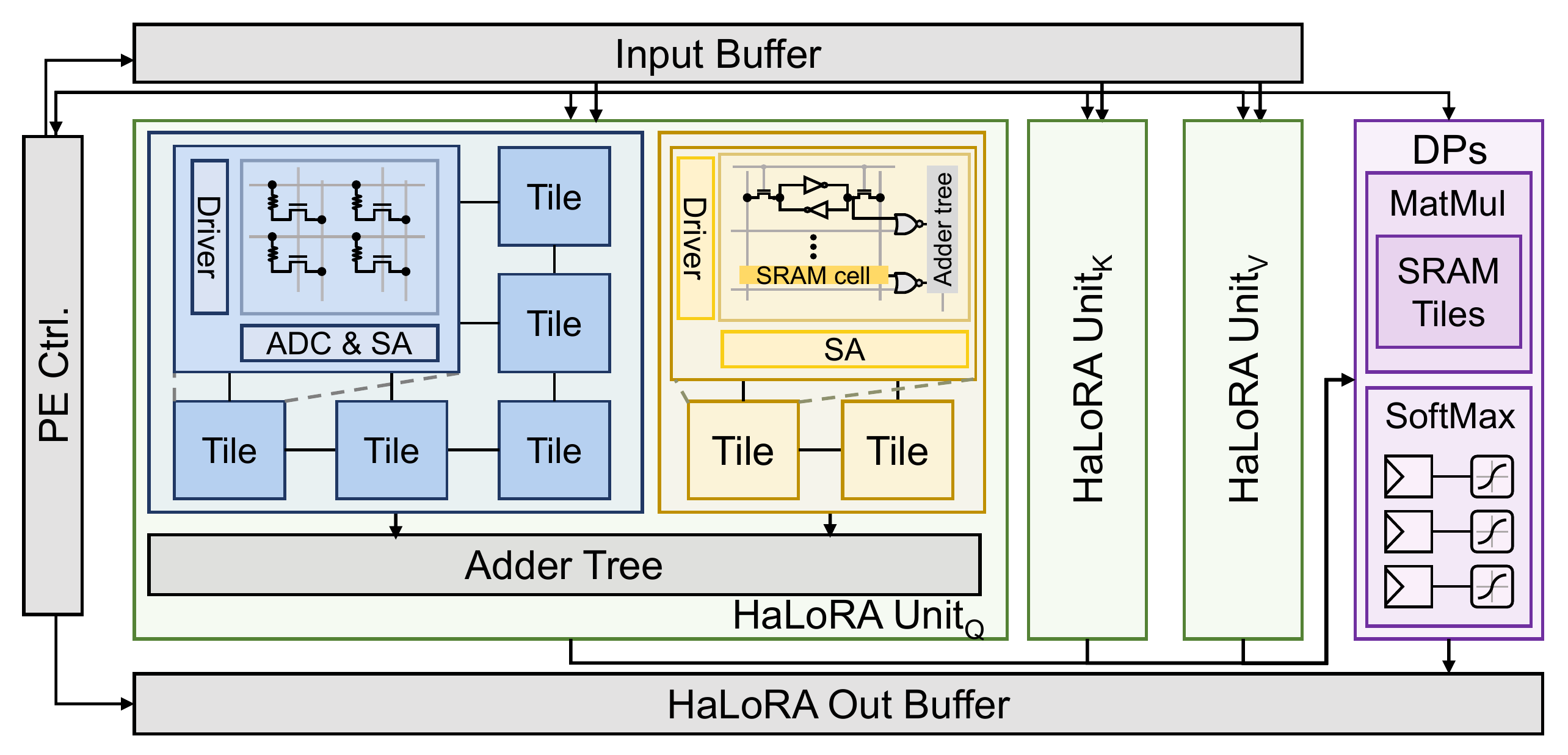}
\caption{
\re{The designed hybrid CIM architecture for attention blocks in HaLoRA. This architecture comprises the main processing elements (PEs) named HaLoRA units, data buffer, controller (PE ctrl.), adder tree, and dynamic processors (DPs) which handle MatMul and SoftMax operations. The HaLoRA unit integrates an RRAM-based analog CIM module and an SRAM-based digital CIM module, both composed of multiple tiles. RRAM tiles consist of RRAM arrays and peripheral circuits including drivers, analog-to-digital converters (ADCs), and shift adders (SAs), while each SRAM tiles comprise SRAM arrays, drivers, adder trees and SAs.}
}
\label{arch_fig}
\end{figure*}
% For attention blocks, while previous works \cite{ReTransformer,ReTransformer_2} demonstrate fully RRAM-based deployment, the dynamic matrix-matrix multiplication (MatMul) requires extensive write-verify operations during inference. To achieve efficient dynamic matrix operations, the MatMul modules are allocated to SRAM for lower memory writing overhead.
For attention blocks, dynamic matrix-matrix multiplication (MatMul) necessitates extensive write-verify operations in RRAM \cite{ReTransformer,ReTransformer_2}. To mitigate memory write overhead, we allocate MatMul modules to SRAM-based CIM. \re{The detailed architecture and dataflow for the attention block are presented in Fig. \ref{arch_fig}. This architecture comprises the main processing element (PE) named HaLoRA unit, data buffer, controller, adder tree, and dynamic processor (DP) which handles MatMul and SoftMax operations. The HaLoRA unit concurrently employs a 1T1R-RRAM-CIM-based analog computing module and a 10T-SRAM-CIM-based digital computing module, both structured as multi-tile arrays. Specifically, the RRAM tiles incorporate drivers for array regulation, analog-to-digital converters (ADCs) for signal readout, and shift adders (SAs) for finalizing MAC operations. Each SRAM tile comprises SRAM-based CIM cells (utilizing 6T-SRAM and 4T-NOR logic) and an adder tree that accumulates 1-bit cell outputs in the digital domain, yielding the final output via shift-add operations. Crucially, these modules facilitate the parallel processing of the backbone and the LoRA branch.}
The output of backbone and the LoRA branch will take a subsequent summation in an adder tree to obtain the Q/K/V matrices. After that, the computed Q/K/V matrices are forwarded to the SRAM-CIM-based MatMul units for processing, followed by SoftMax computation in the digital domain. Throughout this process, the Q, K, and V matrices can be processed in parallel and form an efficient pipeline across consecutive inference tasks. Similarly, the FFN in the Transformer block is processed using multiple HaLoRA units with correlated array scales.

By leveraging the hybrid CIM strategy, the HaLoRA unit fully exploits the density advantages of RRAM and the dynamic processing speed advantages of SRAM CIM, thereby enabling energy-efficient inference while maintaining high accuracy and minimizing the overhead associated with dynamic computations.

\re{
The selection of RRAM and SRAM for our hybrid architecture is driven by their complementary performance and integration advantages.
Compared to other non-volatile options like NAND, RRAM offers superior read/write endurance and energy efficiency~\cite{liu_survey}. 
Similarly, SRAM provides higher operational speed and eliminates the refresh overhead associated with DRAM. 
Crucially, the RRAM-SRAM pair exhibits high compatibility with advanced CMOS logic processes, facilitating the integration of memory and logic on a single chip—a prerequisite for efficient Transformer inference.
}

\subsection{RRAM Noise Modeling}
\label{section:noise}

During the inference stage, hardware non-ideality primarily manifests as random noise arising from device variability \cite{neurosim_inf}. 
Empirical evidence from previous research has demonstrated that such noise can be safely modeled as zero-mean Gaussian noise \cite{yiyu_swim,yiyu_ne_fair,rram_modeling}.
Nevertheless, we also provide the results of the stuck-at faults, and the observations are consistent.
While computations of LoRA branches on SRAM-CIM maintain high precision due to digital computing, noise effects on the LLM backbone executed on RRAM-CIM require special consideration.

% Without loss of generality, we implement a \textbf{block-wise} linear mapping strategy that aligns with practical deployment workflows \cite{neurosim_inf}.
% This strategy introduces noise into the weights of individual sub-blocks to accurately simulate the impact of analog computing noise.
\re{
Without loss of generality, we implement a block-wise noise injection scheme that aligns with practical hardware non-idealities. 
This modeling approach partitions the weight matrix into individual sub-blocks and applies the statistical noise model to each, accurately simulating the impact of analog computing noise on specific hardware tiles~\cite{neurosim_inf}. 
By using this method, the simulation results reflect the practical constraints and noise characteristics of the hybrid architecture during the inference stage.
}
For the target RRAM tiles in the shape of $m\times n$, the magnitude of the simulated \textit{Gaussian} noise is determined by the device-specific standard deviation $\sigma$. 
Thus, the non-ideality simulation of RRAM-CIM macros acting on weight matrix $\mathbf{W}_0 \in \mathbb{R}^{d_1\times d_2}$ can be formulated as:
\begin{equation}
    \{\mathbf{W}_{0,[i,j]}\}_{i=1,j=1}^{k,t} = \operatorname{split}(\mathbf{W}_0), \ k=\lceil\frac{d_1}{m}\rceil, \ t=\lceil\frac{d_2}{n}\rceil,
\end{equation}
\begin{equation}
\begin{aligned}
\mathbf{W}_0^* = \operatorname{cat}(\{\mathbf{W}_{0,[i,j]} + \lambda(\max(\operatorname{abs}(\mathbf{W}_{0,[i,j]})))\}_{i=1,j=1}^{k,t}), \ 
\lambda\sim\mathcal{N}(0,\sigma)
\end{aligned}
\label{eq_rram_noise}
\end{equation}
where $\operatorname{split}(\cdot)$ and $\operatorname{cat}(\cdot)$ denote the matrix splitting and concatenation operations respectively, $\mathbf{W}_{0,[i,j]}$ represents the partitioned weight block, and $\mathbf{W}_0^*$ corresponds to the noise-injected weight matrix.
The $\sigma$ is the noise level, and a larger value denotes more severe noise.

\re{
The Gaussian noise model is a well-established proxy for representing the aggregate non-idealities in RRAM, including programming variations and read noise.
To ensure our simulation aligns with state-of-the-art hardware, we correlate the noise level $\sigma$ with the relative cell-level computing error $1\sigma/\mu$.
For a typical LLM weight distribution (e.g., LLaMA-3.2-1B), the ratio $\max|W|/\text{mean}|W|$ ranges from 5.23 to 9.67. 
Thus, our evaluated range of $\sigma \in [0, 0.02]$ corresponds to a variation level up to approximately 19.34\%, covering both optimized memristive engines ($\sim$2.54\%) and unoptimized systems ($\sim$16.1\%)~\cite{wang2025near}.
}
Moreover, we also report the results of blockwise \textit{Stuck-at Faults} in Section \ref{sec-sp-noise}.
These approaches ensure simulation results that reflect the practical constraints and capabilities of the hybrid architecture.

% \re{
% Our block-wise noise injection strategy is specifically designed to align with practical quantization and deployment workflows for large-scale weights.
% In linear quantization, weight elements are mapped to the conductance range of RRAM devices, where the maximum absolute value within a block determines the mapping scale. 
% While the physical device noise magnitude remains consistent, the actual impact on numerical representation depends on this block-specific scale. 
% Within each CIM macro, noise from individual devices is naturally accumulated through analog multiply-accumulate (MAC) operations. 
% Outside the macro, since the analog outputs are converted back to the digital domain via ADCs, subsequent operations such as shift-add and data transmission are performed with digital precision.
% Therefore, the influence of noise error during post-macro transmission is considered negligible in our model.
% }

\subsection{Hardware-aware Low-Rank Adaptation~(HaLoRA)}
\label{sec:halora}

In the hybrid CIM architecture, the inherent noise from RRAM would lead to performance degradation.
Fig. \ref{intro_fig} indicates an example of the side effect during inference with such noise.
Furthermore, as shown in Table \ref{table_main}, the noise at the level of $\sigma=0.02$ would decrease the average scores by 21.7 and 15.8 for the LLaMA-3.2 1B and 3B models, respectively.

To this end, we propose a novel HaLoRA method to train a robust LoRA branch.
Our key insight is to minimize the gap between the training trajectories under both ideal and noisy conditions during the training process, and thus get a better LoRA branch that is robust to the noise and accurate during inference.
The detailed process to model this gap is as follows.

Considering the gradients for matrices $\mathbf{A}$ and $\mathbf{B}$ in LoRA, we have:
\begin{equation}
\label{eq_grad_ori1}
\frac{\partial \mathcal{L}}{\partial \mathbf{A}} = \frac{\partial \mathcal{L}}{\partial \mathbf{W}} \mathbf{B}^T
\end{equation}
\begin{equation}
\label{eq_grad_ori2}
\frac{\partial \mathcal{L}}{\partial \mathbf{B}} = \mathbf{A}^T \frac{\partial \mathcal{L}}{\partial \mathbf{W}}, 
\end{equation}
where $\mathbf{W}$ is the merged weight defined in Equation \ref{eq_lora}. 
For the ideal condition without noise, the updated LoRA branch can be formulated as:
\begin{equation}
% \small
\label{eq_ori_update}
\begin{aligned}
    & \ \ \ \ (\mathbf{A} - \eta \frac{\partial \mathcal{L}}{\partial \mathbf{A}})
    (\mathbf{B} - \eta \frac{\partial \mathcal{L}}{\partial \mathbf{B}}) \\
    & = (\mathbf{A} - \eta  \frac{\partial \mathcal{L}}{\partial \mathbf{W}} \mathbf{B}^T)
    (\mathbf{B} - \eta \mathbf{A}^T \frac{\partial \mathcal{L}}{\partial \mathbf{W}}) \\
    &\approx \mathbf{A}\mathbf{B}-\eta \mathbf{A} \mathbf{A}^T \frac{\partial \mathcal{L}}{\partial \mathbf{W}} -  \eta  \frac{\partial \mathcal{L}}{\partial \mathbf{W}} \mathbf{B}^T \mathbf{B} ,
\end{aligned}
\end{equation} 
where $\eta$ is the learning rate such as 1e-3 and thus we discard the item $\eta^2 \frac{\partial \mathcal{L}}{\partial \mathbf{W}} \mathbf{B}^T \mathbf{A}^T \frac{\partial \mathcal{L}}{\partial \mathbf{W}}$.
Considering the noise existed in $\mathbf{W}_0$, we have:
\begin{equation}
    \mathbf{W}^{*}=\mathbf{W}_0^*+ \Delta \mathbf{W}.
\end{equation}
Similarly, we can thus get the updated LoRA branch under the same initialization with the noise:
\begin{equation}
% \small
\label{eq_noisy_update}
\begin{aligned}
    & \ \ \ \ (\mathbf{A} - \eta \frac{\partial \mathcal{L}}{\partial \mathbf{A}})
    (\mathbf{B} - \eta \frac{\partial \mathcal{L}}{\partial \mathbf{B}}) \\ 
    &= 
    (\mathbf{A} - \eta  \frac{\partial \mathcal{L}}{\partial \mathbf{W^*}} \mathbf{B}^T)
    (\mathbf{B} - \eta \mathbf{A}^T \frac{\partial \mathcal{L}}{\partial \mathbf{W^*}}) \\
    &\approx \mathbf{A}\mathbf{B}-\eta \mathbf{A} \mathbf{A}^T \frac{\partial \mathcal{L}}{\partial \mathbf{W^*}} -  \eta  \frac{\partial \mathcal{L}}{\partial \mathbf{W^*}} \mathbf{B}^T \mathbf{B} .
\end{aligned}
\end{equation}

\input{figures_tables/Table_algo}

To align the optimization process under ideal and noisy conditions, we aim to minimize the gap $\delta$ between Equation \ref{eq_ori_update} and Equation \ref{eq_noisy_update}:
\begin{equation}
\begin{aligned}
    \min \delta &= ||(\mathbf{A}\mathbf{B}-\eta \mathbf{A} \mathbf{A}^T \frac{\partial \mathcal{L}}{\partial \mathbf{W}} -  \eta  \frac{\partial \mathcal{L}}{\partial \mathbf{W}} \mathbf{B}^T \mathbf{B}) - (\mathbf{A}\mathbf{B}-\eta \mathbf{A} \mathbf{A}^T \frac{\partial \mathcal{L}}{\partial \mathbf{W^*}} -  \eta  \frac{\partial \mathcal{L}}{\partial \mathbf{W^*}} \mathbf{B}^T \mathbf{B})|| \\
    &= \eta || \mathbf{A} \mathbf{A}^T ( \frac{\partial \mathcal{L}}{\partial \mathbf{W^*}}- \frac{\partial \mathcal{L}}{\partial \mathbf{W}}) + (\frac{\partial \mathcal{L}}{\partial \mathbf{W^*}}-\frac{\partial \mathcal{L}}{\partial \mathbf{W}}) \mathbf{B}^T \mathbf{B} ||.
\end{aligned}
\end{equation}
The learning rate $\eta$ is fixed and can be discarded.
Using $||\mathbf{X}\mathbf{Y}|| \leq ||\mathbf{X}|| ||\mathbf{Y}||$ and $||\mathbf{X}+\mathbf{Y}|| \leq ||\mathbf{X}||+ ||\mathbf{Y}||$, we can instead optimize the upper bound $\delta^*$ of $\delta$:
\begin{equation}
% \small
\label{eq11}
\begin{aligned}
    \min \delta^* & = 
    ||\mathbf{A} \mathbf{A}^T|| \ ||( \frac{\partial \mathcal{L}}{\partial \mathbf{W^*}}- \frac{\partial \mathcal{L}}{\partial \mathbf{W}})|| +  || (\frac{\partial \mathcal{L}}{\partial \mathbf{W^*}}-\frac{\partial \mathcal{L}}{\partial \mathbf{W}})|| \ ||\mathbf{B}^T \mathbf{B}||  \\
    &= || (\frac{\partial \mathcal{L}}{\partial \mathbf{W^*}}-\frac{\partial \mathcal{L}}{\partial \mathbf{W}})|| \ (||\mathbf{A} \mathbf{A}^T|| + ||\mathbf{B}^T \mathbf{B}||).
\end{aligned}
\end{equation}
% Since noise is stochastic and within a scope, then we can simply the optimization target:
Fully minimizing the upper bound $\sigma^*$ in Equation \ref{eq11} is intractable. 
The gradient difference term, $|| (\frac{\partial \mathcal{L}}{\partial \mathbf{W^*}}-\frac{\partial \mathcal{L}}{\partial \mathbf{W}})||$ is not a constant.
Instead, it is a complex function that depends on the input data, the stochastically sampled noise, and the LoRA parameters $\mathbf{A}$ and $\mathbf{B}$ themselves.

Therefore, rather than attempting to optimize this intractable and data-dependent sensitivity term, we adopt a theoretically-inspired and pragmatic approach.
We shift our focus to minimizing the other component of the product in $\delta^{*}$: the structural term $||\mathbf{A} \mathbf{A}^T|| + ||\mathbf{B}^T \mathbf{B}||$.
% This simplification reframes the goal into a structural optimization problem. 
% Our key insight is that this term is not merely a simplification, but a powerful structural regularizer that constrains the LoRA matrices. 
This term effectively measures the self-correlation of the row vectors in matrix $\mathbf{A}$ and the column vectors in matrix $\mathbf{B}$, respectively.
Minimizing these norms encourages the row and column vectors to be more orthogonal to each other.
% This induced orthogonality promotes a more uniform distribution of representational information across the low-rank subspace, rather than concentrating it in a few dominant directions. 
Consequently, the output becomes inherently less sensitive to perturbations in the noisy RRAM weights, as any single-directional noise is diluted.
In summary, minimizing this structural term acts as a pragmatic proxy objective to reduce the overall upper bound $\delta^{*}$. 
% It guides the optimization to find a "stabler" solution that is fundamentally more robust to the hardware noise. 
Hence, the optimization objective is:
% Therefore, rather than attempting to optimize this intractable and data-dependent sensitivity term, we adopt a theoretically-inspired, pragmatic approach. 
% We focus on minimizing the other component of the product in $\sigma^*$: the structural term $||\mathbf{A} \mathbf{A}^T|| + ||\mathbf{B}^T \mathbf{B}||$.
% This term is data-agnostic, easily computable, and directly controllable during training. 
% Our key insight is that by explicitly minimizing this structural term, we are imposing a regularization that constrains the LoRA matrices. 
% This serves as a proxy objective to reduce the overall upper bound $\sigma^*$. 
% This simplification reframes our goal into a structural optimization problem:
\begin{equation}
\label{eq_final_goal}
    \min \delta^* \longrightarrow \min ||\mathbf{A} \mathbf{A}^T|| + ||\mathbf{B}^T \mathbf{B}||.
\end{equation}
Finally, the goal is simplified into Equation \ref{eq_final_goal}, which is \textit{agnostic} to noise.
In HaLoRA, we select the Euclidean norm.
As shown in Fig. \ref{method_fig}(a), the training loss to update the LoRA branch is:
\begin{equation}
    \mathcal{L}_{total} = \mathcal{L} + \mu \mathcal{L}_{reg},
\end{equation}
where $\mu$ is the hyperparameter for loss weight and $\mathcal{L}_{reg}=||\mathbf{A} \mathbf{A}^T||_2+||\mathbf{B}^T \mathbf{B}||_2$.
Algorithm \ref{alg1} indicates the details of the proposed HaLoRA.

The term $||\mathbf{A} \mathbf{A}^T|| + ||\mathbf{B}^T \mathbf{B}||$ measures the self-correlation of the row vectors in matrix $\mathbf{A}$ and the column vectors in matrix $\mathbf{B}$, respectively. 
Minimizing these norms effectively encourages the row vectors and the column vectors to be more orthogonal to each other. 
This orthogonality promotes a more uniform distribution of representational information across the low-rank subspace, rather than concentrating it in a few dominant directions. 
Consequently, the model's output becomes less sensitive to perturbations in the noisy RRAM weights, as any single-directional noise is \textit{diluted} and its impact on the final prediction is significantly \textit{attenuated}. 

\re{
It is worth noting that our proposed loss function $\mathcal{L}_{reg}$ is designed as a structural regularizer that is independent of specific input instances or transient noise values.
While this formulation does not explicitly model input-dependent non-idealities such as quantization errors or analog-to-digital converter (ADC) thermal noise, it provides a principled upper bound on noise sensitivity for the low-rank subspace.
For quantization errors, there are some algorithm-level solutions, such as AWQ~\citep{lin2024awq} and APTQ~\citep{guan2024aptq}.
In practical CIM deployment, input-related errors are typically managed at the circuit or architecture level using techniques like dynamic range scaling or error-compensating ADCs \citep{wan2022compute}.
Importantly, our training-stage approach is orthogonal to these deployment-side mitigations. 
By stabilizing the weight matrix structure during finetuning, HaLoRA ensures a robust foundation that enhances the effectiveness of any subsequent hardware-specific optimizations.
}

%% file: figures_tables/Table_algo.tex
\begin{algorithm}[!t]
    % \small
% 	\vskip -0.03in
	\caption{HaLoRA}
	\label{alg1}
	
	{\bf Input:} Train data $\mathcal{D}_{train}$, Finetuned LLM $M$
	
	{\bf Params:} $N$: steps to apply alignment, $\mu$: loss weight for regularization loss

	{\bf Output: } Updated LoRA branch.
	
	\begin{algorithmic}[1]
        \STATE Insert LoRA for selected linear layer in $M$
	    \STATE Initialize LoRA branch where $\mathbf{A}$ as Kaiming Uniform distribution and $\mathbf{B}$ as Zero matrix.
	    % \STATE $k \gets 0$ ; ${M} \gets \left[ \ \right]$
        \STATE Sample Noise to pretrained weight $\mathbf{W}_{0}$ following Equation \ref{eq_rram_noise} 
		\FOR{$i=0$ to max training steps}
        \STATE Forward a batch from $\mathcal{D}_{train}$ through $M$, derive the gradients $g_{ori}$ for LoRA to update following Equation \ref{eq_grad_ori1} and \ref{eq_grad_ori2}
		\IF{ $i \% N == 0$} 
        \STATE Sample Noise to pretrained weight $\mathbf{W}_{0}$ following Equation \ref{eq_rram_noise}  
		\STATE Calculate regularization loss $\mathcal{L}_{reg}$ following Equation \ref{eq_final_goal}
        \STATE backward loss to get the regularization gradient $g_{reg}$
        \STATE $g_{ori} \gets g_{ori} + \mu g_{reg}$
		\ENDIF
		\STATE Update the LoRA branch with corresponding $g_{ori}$ while keeping $\mathbf{W_0}$ frozen
		\ENDFOR
		\STATE {\bf return} Optimized the LoRA branch
	\end{algorithmic}
\end{algorithm} 

%% file: section/4-Experiments.tex
\section{Experiments and Analysis}

In this section, we conducted a comprehensive comparison between vanilla LoRA and HaLoRA finetuning popular open-source LLaMA and Qwen models across various benchmarks under different noise settings.
Moreover, we perform the ablation study on the hyperparameter $\mu$, compare the hardware simulation results, and conduct a case study on the generated context.

\subsection{Experimental Setup}
\label{sec:Experimental_setup}
\subsubsection*{Non-ideal effect simulation} 
In the hybrid CIM architecture, we consider the inference results based on digital SRAM macros to be relatively accurate. 
\re{For the analog RRAM macros, based on noise levels reported in published RRAM chip studies \cite{yao_nature,device_2}, the standard deviation of the injected Gaussian noise can be set within the range of 0.01 to 0.02.}
To comprehensively evaluate practical noise mitigation strategies, including redundant mapping and bit-splitting techniques, we extended our validation to encompass lower noise levels ($\sigma =$ 0.005, 0.01, and 0.02). 
Additionally, following the block-wise linear mapping characteristics of weights on physical RRAM crossbars \cite{neurosim_inf}, we partitioned the corresponding weights into 64×64 blocks to align with conventional memory tile dimensions. These blocks were then subjected to noise injection procedures (detailed in Section~\ref{section:noise}).
In addition, we also report the results of considering stuck-at faults.

\subsubsection*{HaLoRA finetuning} We conduct experiments on Qwen2.5 0.5B, which has 0.5 billion parameters, and two variants of the LLaMA-3.2 family: LLaMA-1B and LLaMA-3B, with 1.3 billion and 3.2 billion parameters, respectively. 
Following \cite{hu_llmadapter}, we employ the 170k samples for training.
As shown in Fig. \ref{method_fig}, we insert the LoRA branches into five modules: query/key/value/up/down projection.
\re{
Specifically, the random noise is injected exclusively into the frozen pretrained weights $W_0$ during the forward pass to simulate hardware non-idealities.
The training dataset $\mathcal{D}_{train}$ remains noise-free and clean. 
This approach ensures that the LoRA branches learn to compensate for weight perturbations rather than adapting to noisy inputs.
}
The rank in LoRA is 4.
For the random noise during training, we sample noise from a Gaussian distribution every 400 steps ($N=400$), and add the noise to the original weights.
The noise level $\sigma$ is set as the typical value 0.01 following  \citep{agarwal2016resistive}.
We train on a single NVIDIA A100 GPU with a batch size of 16 for 3 epochs while the learning rate is 1e-4.
The $\mu$ is set to be 0.1.
It takes around 2 hours to finetune the 1B version LLaMA-3.2.

\input{figures_tables/Table_dataset}

\subsubsection*{Evaluation}
% \textcolor{red}{details}
We evaluate finetuned models on 6 popular benchmarks, such as 1) ARC-e (Easy set of AI2 Reasoning Challenge), 2) OBQA (OpenBook Question Answering), 3) SIQA (Social Interaction QA), 4) ARC-c (Challenge set of AI2 Reasoning Challenge), 5) WinoG. (Winograd Schema Challenge), and 6) PIQA (Physical Interaction QA).
% We refer the readers to~\cite{wu_moslora} for more details about these benchmarks. 
Table \ref{tab-dataset} indicates the details of these datasets.
The difficulty and subject of the questions in these benchmarks vary and thus better to perform a comprehensive evaluation of the finetuned model.
During inference, we sample the noise 5 times under the random seed \{1,2,3,4,5\} for each noise level from \{0.005, 0.01, 0.02\}. Meanwhile, we also report the performance without noise.
For all the benchmarks, both the average and standard deviation of accuracy are reported.

%%%%%%%%%%%%%%%%%%%%%%%%%%%%%
%%%%%%%%%%%%%%%%%%%%%%%%%%%%%
\subsection{Main Results}

\input{figures_tables/Table_main_results}

Table~\ref{table_main} presents the results on 6 benchmarks when finetuning Qwen2.5 0.5B and LLaMA-3.2 1B/3B.
For each benchmark, we report the noise-free evaluation results and the average with standard deviation for five runs when adding noise under the seeds 1-5.
Besides, we also report the overall average of these average scores on 6 benchmarks.
Some key findings can be summarized as:
\begin{itemize}
    \item \textbf{HaLoRA demonstrates better performance at noise-free setting.} As LLMs inherently tend to overfit during dataset training \cite{wu_noisytune}, HaLoRA's noise injection effectively enhances the diversity of model representations during the finetuning process, leading to improved performance on test sets. 
    Specifically, under noise-free deployment conditions, HaLoRA improves average performance by 3.0/5.3/0.6 points compared to vanilla LoRA on Qwen2.5 0.5B/LLaMA-3.2 1B/3B models, respectively.
    Moreover, such noise would lead to performance degeneration, and a higher noise level results in a more significant drop.
    \item \textbf{HaLoRA demonstrates superior accuracy at all noise levels.} Compared to vanilla LoRA, HaLoRA demonstrates enhanced performance across all six datasets consistently.
    Notably, at a deployment noise level of 0.02, HaLoRA achieves substantial performance improvements over vanilla LoRA: 20.5 points for Qwen 0.5B (48.6 vs. 28.1), 22.7 points (63.1 vs. 40.4) for LLaMA-3.2 1B, and 13.5 points (78.4 vs. 64.9) for LLaMA-3.2 3B. 
    Furthermore, HaLoRA shows significantly reduced performance degradation as noise levels increase. 
    When noise levels escalate to 0.02, HaLoRA's average performance degradation is only 21\% (4.5 vs. 21.9 points) and 18\% (2.9 vs. 15.8 points) of vanilla LoRA's degradation for LLaMA-3.2 1B and 3B, respectively. 
    \item \textbf{HaLoRA demonstrates superior stability under noisy conditions.} At identical noise levels, HaLoRA exhibits significantly lower performance variance across different noise directions. For LLaMA 1B, across five experimental runs, HaLoRA's performance variance on WinoG. and PIQA datasets are merely 7\% (1.0 vs. 12.9) and 10\% (1.7 vs. 17.3) of vanilla LoRA's variance, respectively. Similarly, for LLaMA 3B, HaLoRA achieves remarkably lower variance on ARC-e and WinoG. datasets, showing only 3\% (0.3 vs. 9.2) and 11\% (0.9 vs. 8.4) of vanilla LoRA's variance.
    \item \textbf{Larger model shows better robustness against noise.} 
    Comparing the performance of LLaMA-3.2 1B and 3B across multiple datasets under identical noise levels, we can find that both vanilla LoRA and HaLoRA (particularly the latter) exhibit reduced accuracy degradation in larger models (accuracy drops of 15.8 vs. 21.9 for vanilla LoRA, and 2.9 vs. 4.5 for HaLoRA). 
    Additionally, larger models show more stable performance, as evidenced by lower standard deviations (maximum values of 5.5 vs. 17.3 for vanilla LoRA, and 1.5 vs. 1.9 for HaLoRA). 
    These results suggest that the model scale positively correlates with noise resilience.
\end{itemize}

In summary, HaLoRA outperforms LoRA for all the benchmarks under various noise levels, providing a robust and effective solution to the noise from RRAM when deploying finetuned LLMs on hybrid CIM architectures.
Our findings also reveal that larger models (especially with HaLoRA) tend to demonstrate better tolerance to hardware-induced noise compared to the smaller models.
Nevertheless, one interesting finding is that noise at the training stage would improve the noise-free performance at the inference stage.

% \subsection{Further Analysis}

% \begin{figure}[t]
%     \centering
%     \begin{subfigure}[t]{\linewidth}
%     \caption{OBQA}
%     \includegraphics[width=\linewidth]{figures_tables/OBQA.pdf}
%     % \vskip -0.1in
%     \label{fig:OBQA}
%     \end{subfigure}
%     \vskip -0.2in
%     \begin{subfigure}[t]{\linewidth}
%     \caption{SIQA}
%     \includegraphics[width=\linewidth]{figures_tables/SIQA.pdf}
%     % \vskip -0.1in
%     \label{fig:SIQA}
%     \end{subfigure}
%     % \vskip -0.1in
%     \caption{The performance of HaLoRA with different values of $\mu$ and vanilla LoRA on the OBQA and SIQA datasets.}
%     \label{fig:sensitivity}
% \end{figure}

\subsection{Computing Overhead}

\input{figures_tables/Table_compute_cost}

In the proposed HaLoRA, we design an extra loss to minimize the gap between the optimization trajectories of the LoRA branch under both ideal and noisy conditions.
During the inference stage, the computing overhead is exactly the same as vanilla LoRA.
Therefore, we further compare the training overhead.
Table \ref{tab-cost} reports the results and performance of LoRA and HaLoRA when finetuning the Qwen2.5 0.5B model on a single Nvidia A100 GPU.
We can find that the HaLoRA brings an extra training time of 0.12h (1.95 vs. 1.83) and extra GPU memory of 0.9G (19.7 vs. 18.8).
The extra cost is quite slight during training.
However, during the inference, HaLoRA achieves better performances considering the inherited noises from the RRAM, such as a gain of 20.5 on the average score.
% Therefore, we consider it acceptable since a quite slight extra training cost can bring a large improvement in robustness.
Therefore, we believe that such an extra computing overhead is acceptable since a better finetuned LLM would save more time on downstream tasks.

 \begin{figure*}[t]
    \centering
    \includegraphics[width=\linewidth]{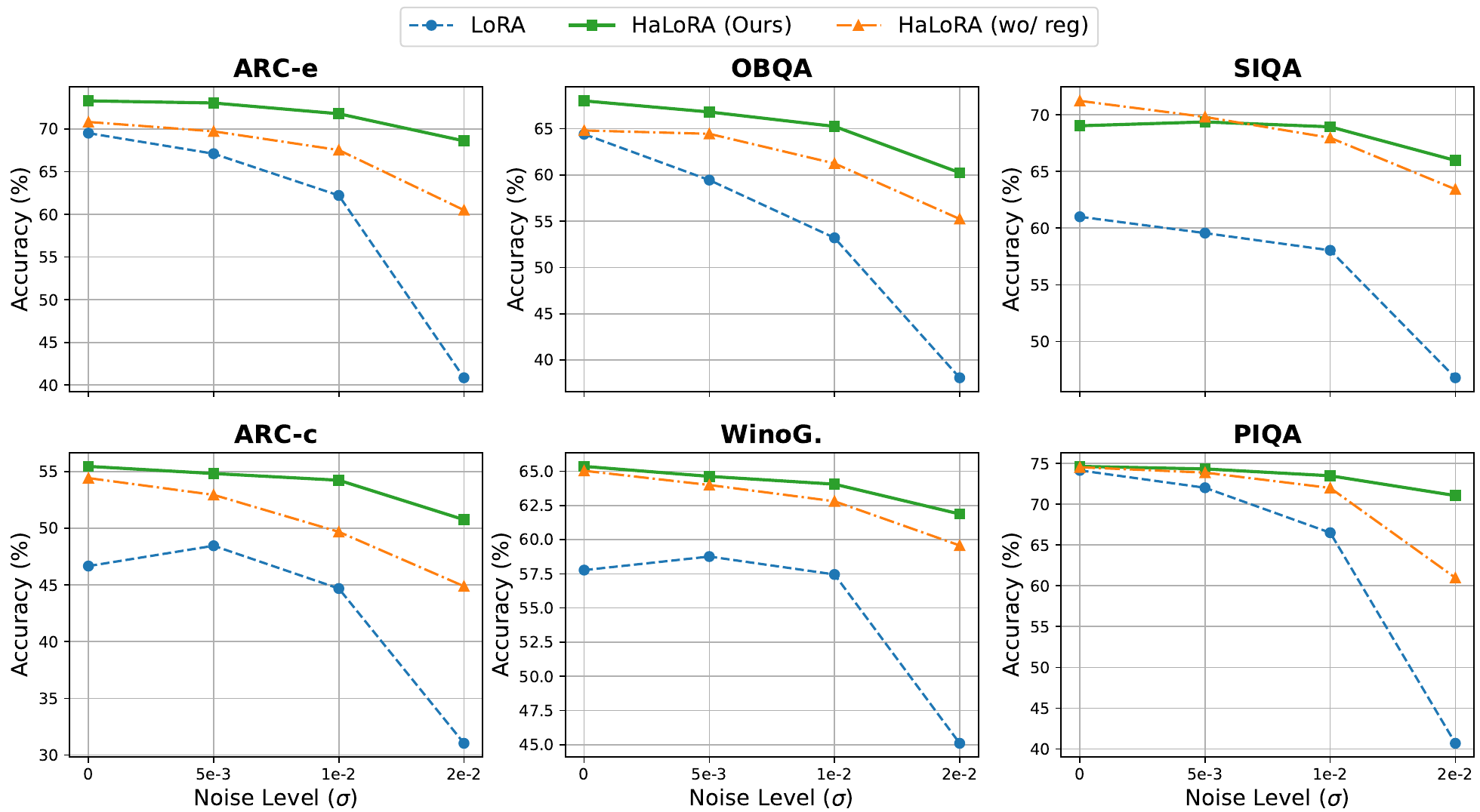}
    \caption{\re{
    Ablation of HaLoRA removing regularization (denoted as HaLoRA (wo/ reg)) when tuning Llama-3.2-1B, i.e., noise-aware training \textbf{only}.
    }
    }
    \label{fig:aba_reg}
\end{figure*}

\subsection{Ablation on Regularization}
\label{sec:aba_reg}

% \re{
% In HaLoRA, we introduce a novel regularization loss $\mathcal{L}_{reg}$ weighted by $\mu$ while injecting noise during training.
% Before analyzing the sensitivity to various $\mu$ values, it is crucial to ablate and quantify the precise contribution of the $\mathcal{L}_{reg}$ term itself.
% To isolate this effect, we compare the baseline LoRA against our full HaLoRA method ($\mu=0.1$) and a strong baseline of $\mu=0$, which represents standard noise-aware training without the structural regularization.
% As shown in Fig. \ref{fig:aba_reg}, the $\mathcal{L}_{reg}$ term provides essential, targeted robustness that simple noise-aware training alone does not capture.
% Specifically, while noise-aware training ($\mu=0$) improves performance over vanilla LoRA, its gain is limited under high noise levels.
% At $\sigma=0.02$, the full HaLoRA ($\mu=0.1$) achieves an average score of 63.1, surpassing the noise-aware baseline ($\mu=0$) by a significant margin of 5.7 points. 
% This result disentangles the benefits of the two components: noise injection helps the model adapt to perturbations, but the structural regularizer ($||AA^T||_2 + ||B^TB||_2$) is the key to stabilizing the low-rank subspace and ensuring long-term robustness.
% }

\re{
In the HaLoRA framework, we introduce a novel regularization loss $\mathcal{L}_{reg}$ weighted by the hyperparameter $\mu$, integrated with a dynamic noise injection mechanism during training.
To isolate and quantify the precise contribution of the structural regularization term to model robustness, we conducted a critical ablation study.
Fig. \ref{fig:aba_reg} illustrates the accuracy trends across six commonsense reasoning benchmarks as the hardware noise level $\sigma$ increases.
The results indicate that simply injecting noise during training ($\mu=0$) allows the model to achieve better robustness than vanilla LoRA under low-intensity hardware noise.
This suggests that the model can learn to adapt to perturbations to a certain degree through passive exposure.
However, pure noise injection shows significant limitations as the noise level $\sigma$ escalates to $0.02$.
Under these extreme conditions, the full HaLoRA method demonstrates a substantial performance advantage.
Specifically, at $\sigma=0.02$, HaLoRA achieves a score of 71.1 on PIQA, surpassing the noise-aware baseline ($\mu=0$) by a significant margin of 10.1 points.
These findings successfully disentangle the unique benefits of the two components: while noise injection helps the model adapt to specific perturbation distributions, the structural regularizer ($||AA^T||_2 + ||B^TB||_2$) is the key to stabilizing the low-rank subspace.
By encouraging orthogonality among the row and column vectors of the LoRA matrices, the regularizer ensures that representational information is distributed more uniformly.
This prevents the model from being overly sensitive to any single direction of perturbation, effectively diluting the impact of RRAM-induced noise.
}

%% 不支持 subfig
 \begin{figure*}[t]
    \centering
    \includegraphics[width=\linewidth]{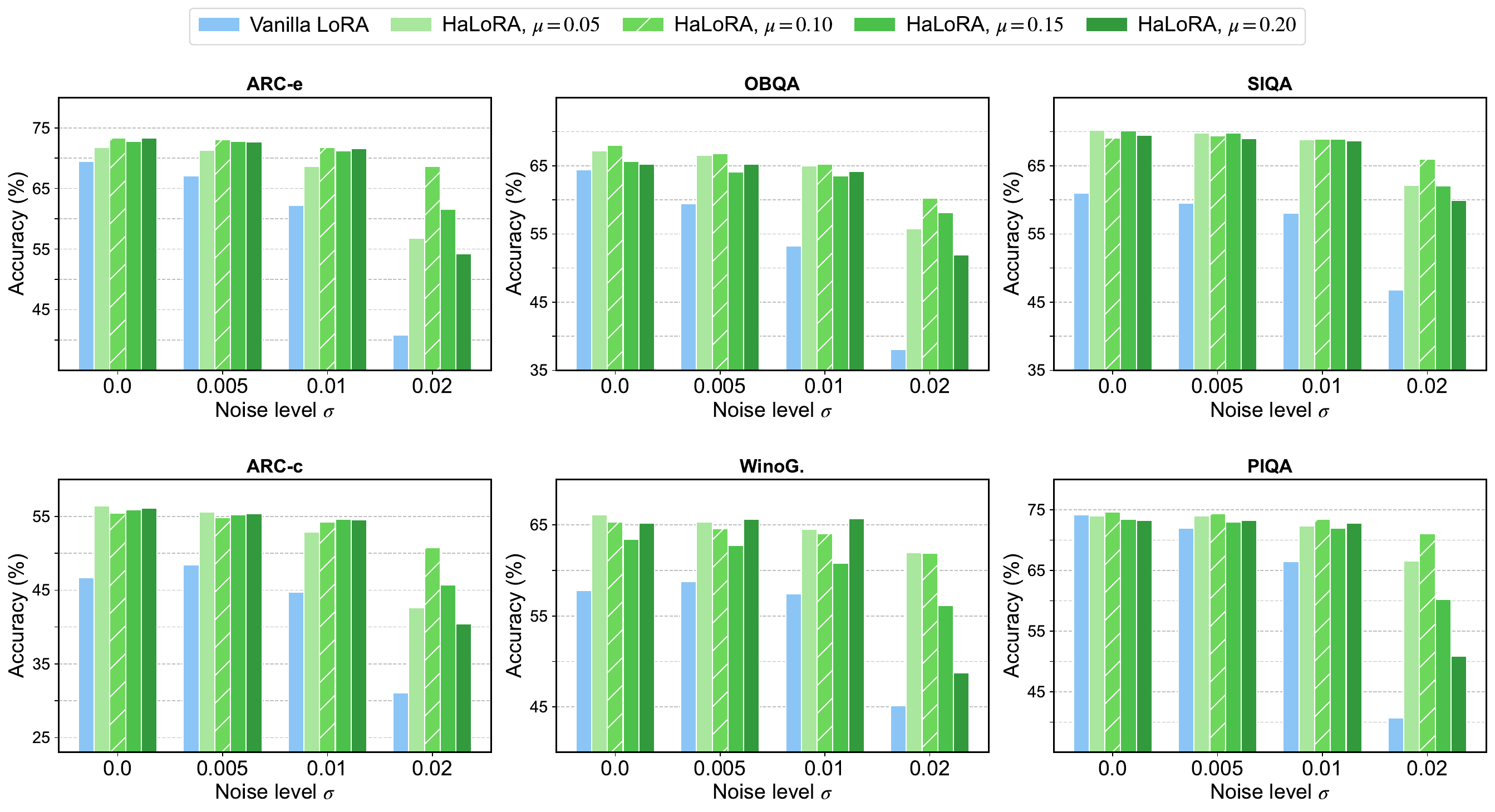}
    \caption{The performance of HaLoRA with different values of $\mu$ and vanilla LoRA on the six commonsense reasoning datasets when finetuning LLaMA-3.2 1B.
    We report the average score of five runs with seeds being 1-5.
    We can find that HaLoRA with different $\mu$ consistently outperforms vanilla LoRA, demonstrating the effectiveness and robustness of the proposed HaLoRA framework. }
    \label{fig:sensitivity}
\end{figure*}

\subsection{Sensitivity of $\mu$}
\label{sec:sensitivity}

% In HaLoRA, we introduce a novel regularization loss $\mathcal{L}_{reg}$ weighted by $\mu$ while injecting noise during training.
% Before analyzing the sensitivity to various $\mu$ values, it is crucial to ablate and quantify the precise contribution of the $\mathcal{L}_{reg}$ term itself.
% To isolate this effect, we compare the baseline LoRA against our full HaLoRA method (which uses $\mu=0.1$ as set in Section 4.1) and a strong baseline of $\mu=0$, which employs the full noise-aware training process \emph{without} the regularization loss.
% As shown in Table \ref{aba1}, $\mathcal{L}_{reg}$ term provides essential, targeted robustness against hardware noise that simple noise-aware training alone does not capture, and its contribution becomes more critical as the noise increases.

To evaluate the sensitivity of HaLoRA on different $\mu$, we further set $\mu$ to be 0.05/0.15/0.20 when finetuning the LLaMA-3.2 1B model.
Fig. \ref{fig:sensitivity} illustrates the average performance scores across different noise levels $\sigma$ and $\mu$ values on all six benchmarks with random seeds of 1 to 5, respectively.
Overall, HaLoRA consistently outperforms baseline LoRA across all six benchmarks for various $\mu$, demonstrating the effectiveness and robustness of HaLoRA.
Specifically, considering the benchmark SIQA while the noise level $\sigma$ is 0.02, HaLoRA surpasses baseline LoRA consistently with improvements of 15.37, 19.18, 15.30, and 13.14 with $\mu$ being 0.05, 0.10, 0.15, and 0.20, respectively.

Moreover, both excessively large and small values of $\mu$ lead to a performance drop, while 0.1 emerges as a suitable trade-off.
A larger $\mu$ imposes stronger regularization on the LoRA branch, leading to a shift between the optimal points of $\mathcal{L} + \mu \mathcal{L}_{reg}$ and $\mathcal{L}$.
Conversely, a smaller $\mu$ reduces the impact of the regularization term, rendering the performance more fragile.
For instance, considering the ARC-e benchmark, HaLoRA ($\mu$=0.20) brings higher results than ($\mu$=0.10) without noise but is quite sensitive to noise at level 0.02.
Furthermore, the regularization weight $\mu$ serves as a critical trade-off between task performance and noise robustness.
We leave exploring optimal $\mu$ for future work, potentially by setting $\mu$ a learnable parameter.

% Overall, HaLoRA demonstrates relatively stable performance across different values of $\mu$. 
% Although the performance variations slightly increase with higher noise levels, the average results from five experiments show that $\mu$ within the range of [0.05, 0.15] 
% %[0.05,0.20]
% lead to maximum performance changes of 4.4 
% %(8.3) 
% and 3.9 
% %(6.04) 
% points on OBQA and SIQA, respectively. Notably, HaLoRA consistently outperforms Vanilla LoRA across all values of $\mu$.

% As $\mu$ increases from smaller to larger values, we observe distinct behavioral patterns. Smaller $\mu$ enhances the LoRA branch's noise tolerance at the finetuning noise level but leads to performance degradation when the noise level exceeds the finetuning setting ($\sigma=0.02>0.005$). Conversely, excessively large $\mu$ leads to overfitting to precise outputs, compromising the network's inherent noise-tolerant capabilities.

% \subsection{Salt-and-pepper Noise}
\subsection{Stuck-at Faults}
\label{sec-sp-noise}
% Stuck-at fault
 \begin{figure*}[t]
    \centering
    \includegraphics[width=\linewidth]{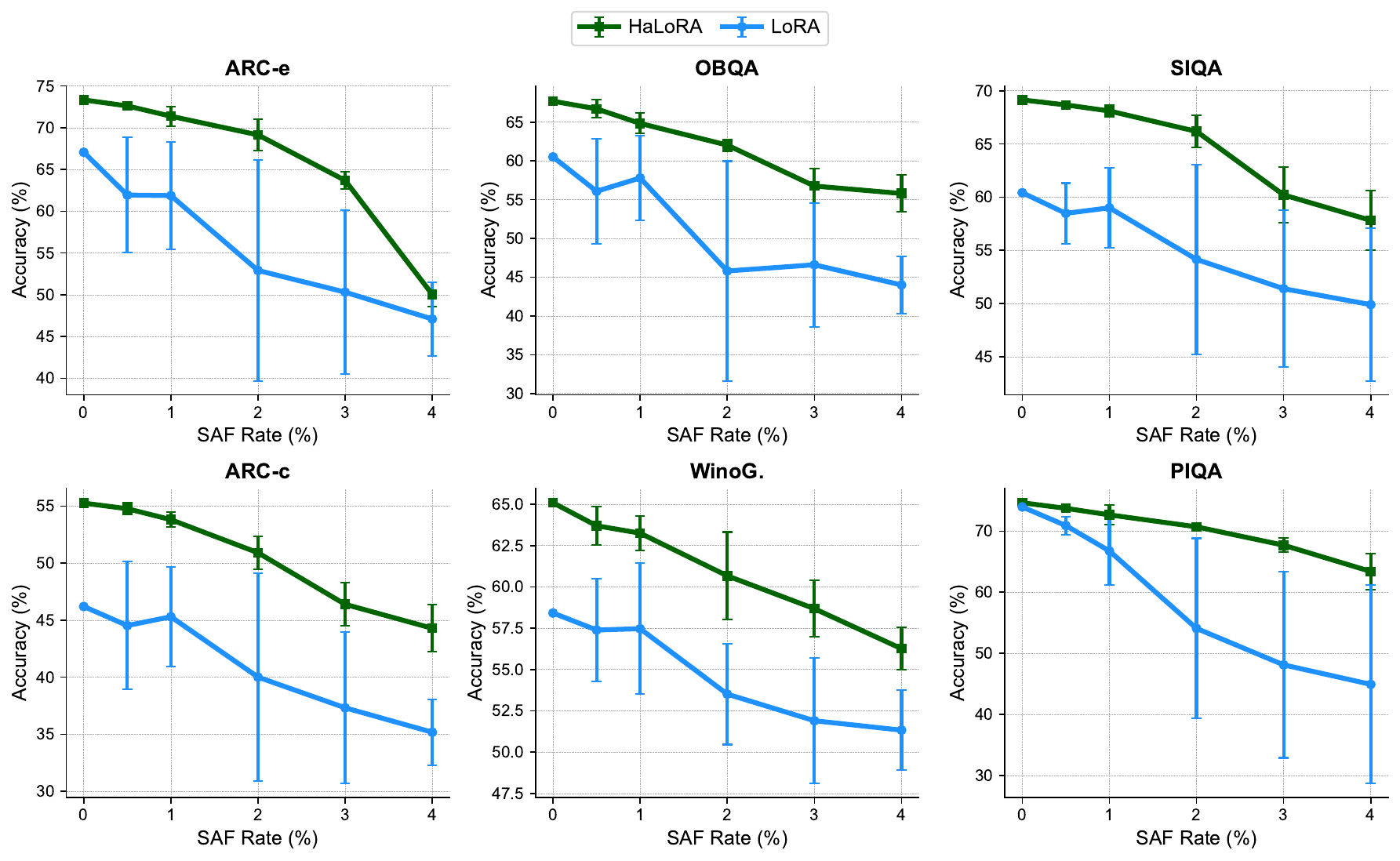}
    \caption{The performance of HaLoRA and LoRA on 6 benchmarks under various stuck-at fault rates on the Qwen2.5 0.5B model.
    For each noise, we repeat 5 times under seeds 1-5 and visualize the average performance with standard deviation.}
    \label{fig:01noise}
\end{figure*}

As mentioned in Section \ref{section:noise}, we model the RRAM noise following a block-wise Gaussian noise, which aligns with practical deployment workflows\cite{neurosim_inf}.
Moreover, we set the noise level to typical values including 5e-3, 1e-2, and 2e-2.
Also, Stuck-at faults are another common type of noise in the RRAM circuit array \cite{sawal2024stuck, xia2017stuck}.
% In this paper, we try to approximate the effect of stuck-at faults by adding salt-and-pepper noise and controlling the stuck-at-fault rates (SAF Rate).
Specifically, we simulate stuck-at faults with different rates to the pretrained weights of the finetuned Qwen model during inference.
After that, we report the average score and standard deviation for five runs under seeds 1-5.

As shown in Fig. \ref{fig:01noise}, we can find that HaLoRA consistently outperforms baseline LoRA across all six benchmarks.
Meanwhile, the standard deviations are far smaller than LoRA (e.g., 1.86 compared to 13.22 on ARC-e with 2\% SAF noise), showing the strong robustness of the proposed HaLoRA.
When the SAF rates equal or less than 3\%, the lower bounds of HaLoRA are even higher than the upper bounds of LoRA, demonstrating the statistical significance of the conclusion that HaLoRA surpasses LoRA.

Considering the performance without noise (i.e., SAF being 0\%), Fig. \ref{fig:01noise} indicates that HaLoRA gets higher scores than LoRA across all six benchmarks, including 6.27 higher on ARC-e (73.33 vs. 67.06), 7.08 on OBQA (67.60 vs. 60.52), 8.72 on SIQA (69.14 vs. 60.42), 9.08 on ARC-c (55.27 vs. 46.19), 6.87 on WinoG. (65.09 vs. 58.42), and 0.68 on PIQA (74.65 vs. 73.97).

Comparing various SAF rates, we can find that a higher SAF rate does not always cause a larger standard deviation, since the importance of each weight is not equal and we add the noise randomly.
Some weights are more important to the final performance than others.
Therefore, adding less noise to the important weights may cause a larger performance drop than adding more noise to the less important weights.
For instance, considering LoRA on the OBQA benchmark, the performance under 0.5\% SAF rate is higher than under 2\%.
Similarly, the standard deviation also follows the same observation that a smaller SAF ratio may lead to a worse result, such as the standard deviations under the 2\% and 4\% SAF rates regarding LoRA on ARC-c.

\re{
In this study, we evaluate HaLoRA using Gaussian noise and stuck-at faults (SAF), which are widely accepted statistical models for characterizing RRAM non-idealities. 
However, we emphasize that the HaLoRA framework is inherently compatible with more complex or device-specific noise distributions. 
In real-world deployment scenarios, our noise-injection training process can be readily adapted to incorporate measured noise signatures from specific hardware batches or individual CIM chips. 
By replacing the generic Gaussian prior with empirical noise features captured from the target device, HaLoRA can leverage the noise-free SRAM-based LoRA branches to perform precise, chip-specific compensation, further bridging the gap between numerical simulation and physical hardware performance.
}

\subsection{Hardware Simulation}
\label{sec:Hardware_simu}

\input{revision/hardware_sim}

\subsection{Case study}

\input{figures_tables/Table_case_study}

We showcase the outputs of LoRA and HaLoRA with and without noise, respectively.
Table \ref{tab_cases} indicates the cases from the ARC-e dataset and corresponding answers of the LoRA-finetuned and HaLoRA-finetuned LLaMA-3.2 1B models.
In this table, we present the input question, ground truth, answers without noise, and answers with Gaussian noise at a level of 2e-2.
We can find that noise would lead to the nonsense output from the LoRA-finetuned LLaMA-3.2 1B, while the HaLoRA-finetuned one shows strong robustness capability and still generates the correct answers.
Such a phenomenon is consistent with Fig. \ref{intro_fig}.
Specifically, LoRA-finetuned LLM can generate the correct answers (i.e., Answer1 for the first case) at a noise-free setting, but fails with noise and generates nonsense output "1/2/3/4".
Meanwhile, the HaLoRA-finetuned model shows superior robustness capability and consistently generates the right output: The correct answer is Answer1.

Moreover, considering the noise-free setting of the second case, HaLoRA-finetuned LLaMA-3.2 1B can generate the correct answers while LoRA-finetuned LLaMA fails, corresponding to the conclusion shown in Table \ref{table_main} that HaLoRA demonstrates better performance under a noise-free setting.
Specifically, HaLoRA gets an average score of 67.6, which is higher than the 62.3 of LoRA. 

%% file: figures_tables/Table_dataset.tex
\begin{table}[!t]
\centering
\caption{The details of six benchmarks for evaluation.}
% \renewcommand{\arraystretch}{1.2}
% \resizebox{\linewidth}{!}
% {
\begin{tabular}{lcp{0.6\linewidth}}
\toprule
\textbf{Dataset} & \textbf{Test Set} & \textbf{Description} \\
\midrule
ARC-e & 1172 & Easy Set of ARC dataset of genuine grade-school level. \\
\midrule
OBQA & 500 & Questions requiring multi-step reasoning, use of additional
commonsense knowledge, and rich text comprehension. \\
\midrule
SIQA & 1954 & Reasoning questions
about people’s actions and their social implications. \\
\midrule
ARC-c & 2376 & Challenge Set of ARC dataset of genuine grade-school level. \\
\midrule
WinoG. & 1267 & Fill-in-a-blank task with binary options to choose the right option for a given sentence, which requires commonsense reasoning. \\
\midrule
PIQA & 1830 & Questions with two
solutions requiring physical commonsense. \\
\bottomrule
\end{tabular}
% }
\label{tab-dataset}
\end{table}

%% file: figures_tables/Table_main_results.tex
\begin{table*}[!t]
\centering
\caption{The comparison between LoRA and HaLoRA on six popular benchmarks when finetuning Qwen2.5 0.5B, LLaMA-3.2 1B, and 3B on commonsense reasoning tasks.
We report the average score and standard deviation of 5 runs when adding noise under seeds 1-5.
HaLoRA outperforms LoRA under the noise-free setting and showcases better robustness at all noise levels.
}
\renewcommand{\arraystretch}{1.2}
\resizebox{\linewidth}{!}
{
\begin{tabular}{c|cc|llllll|c|c}
\toprule
 \multirow{2}{*}{\textbf{Model}} &  \textbf{Training} & \textbf{Inference} & \multirow{2}{*}{\textbf{ARC-e}} & \multirow{2}{*}{\textbf{OBQA}} & \multirow{2}{*}{\textbf{SIQA}} & \multirow{2}{*}{\textbf{ARC-c}} & \multirow{2}{*}{\textbf{WinoG.}} & \multirow{2}{*}{\textbf{PIQA}} & \multirow{2}{*}{\textbf{Avg.}} & \multirow{2}{*}{$\Delta$} \\
 & \textbf{Method} & \textbf{Noise} & & & & & & & &
\\
\midrule
 &  & 0 & 70.5 & 57.0 & 64.8 & 53.2 & 56.0 & 51.4 & 58.8 & - \\
 &  & 5e-3 & 69.1\tiny{$\pm$0.7} & 56.4\tiny{$\pm$1.2} & 59.2\tiny{$\pm$3.8} & 51.8\tiny{$\pm$0.3} & 54.9\tiny{$\pm$2.0} & 48.7\tiny{$\pm$9.2} & 56.7 & - \\
 &  & 1e-2 & 47.5\tiny{$\pm$17.2} & 40.7\tiny{$\pm$12.8} & 48.7\tiny{$\pm$10.5} & 35.9\tiny{$\pm$12.0} & 51.6\tiny{$\pm$1.7} & 44.7\tiny{$\pm$13.4} & 44.9 & - \\
 & \multirow{-3}{*}{LoRA} & 2e-2 & 23.7\tiny{$\pm$19.4} & 18.4\tiny{$\pm$13.1} & 23.5\tiny{$\pm$20.3} & 18.7\tiny{$\pm$15.2} & 46.9\tiny{$\pm$3.2} & 37.2\tiny{$\pm$28.2} & 28.1 & - \\
 \cmidrule{2-11}
 &  & \cellcolor{gg}{0} &  69.5 &  58.2 & 65.0  &  52.9 &  56.6 &  68.7 & \cellcolor{gg}{61.8} & \textcolor{ForestGreen}{+3.0} \\
 &  & \cellcolor{gg}{5e-3} &  68.8\tiny{$\pm$1.1} &  58.3\tiny{$\pm$1.4} &  64.6\tiny{$\pm$0.3} &  52.3\tiny{$\pm$0.5} &  57.4\tiny{$\pm$1.0} &  67.7\tiny{$\pm$0.2} & \cellcolor{gg}{61.5} & \textcolor{ForestGreen}{+4.8} \\
\multirow{-8}{*}{Qwen2.5} &  & \cellcolor{gg}{1e-2} &  66.1\tiny{$\pm$1.4} &  56.3\tiny{$\pm$0.5} &  62.1\tiny{$\pm$1.8} &  49.8\tiny{$\pm$1.8} &  57.2\tiny{$\pm$0.4} &  61.3\tiny{$\pm$5.5} & \cellcolor{gg}{58.8} & \textcolor{ForestGreen}{+13.9} \\
\multirow{-7}{*}{0.5B} & \multirow{-4}{*}{HaLoRA} & \cellcolor{gg}{2e-2} &  52.6\tiny{$\pm$3.9} &  42.7\tiny{$\pm$6.2} &  51.2\tiny{$\pm$8.3} &  40.3\tiny{$\pm$0.6} &  49.1\tiny{$\pm$0.9} &  55.6\tiny{$\pm$3.0} & \cellcolor{gg}{48.6} & \textcolor{ForestGreen}{+20.5} \\
\midrule
 &  & 0 & 69.5 & 64.4 & 61.0 & 46.7 & 57.8 & 74.2 & 62.3 & - \\
 &  & 5e-3 & 67.1\tiny{$\pm$3.2} & 59.4\tiny{$\pm$4.1} & 59.6\tiny{$\pm$0.5} & 48.5\tiny{$\pm$3.4} & 58.8\tiny{$\pm$1.9} & 72.0\tiny{$\pm$1.9} & 60.9 & - \\
 &  & 1e-2 & 62.2\tiny{$\pm$7.6} & 53.2\tiny{$\pm$8.7} & 58.0\tiny{$\pm$1.9} & 44.7\tiny{$\pm$7.4} & 57.5\tiny{$\pm$3.5} & 66.5\tiny{$\pm$4.7} & 57.0 & - \\
 & \multirow{-3}{*}{LoRA} & 2e-2 & 40.9\tiny{$\pm$8.3} & 38.1\tiny{$\pm$6.1} & 46.8\tiny{$\pm$6.4} & 31.0\tiny{$\pm$5.5} & 45.1\tiny{$\pm$12.9} & 40.7\tiny{$\pm$17.3} & 40.4 & - \\
 \cmidrule{2-11}
 &  & \cellcolor{gg}{0} &  73.3 &  68.0 & 69.0  &  55.5 &  65.4 &  74.6 & \cellcolor{gg}{67.6} & \textcolor{ForestGreen}{+5.3} \\
 &  & \cellcolor{gg}{5e-3} &  73.1\tiny{$\pm$0.2} &  66.8\tiny{$\pm$0.4} &  69.4\tiny{$\pm$0.2} &  54.8\tiny{$\pm$0.4} &  64.6\tiny{$\pm$0.6} &  74.3\tiny{$\pm$0.3} & \cellcolor{gg}{67.2} & \textcolor{ForestGreen}{+6.3} \\
\multirow{-8}{*}{LLaMA-3.2} &  & \cellcolor{gg}{1e-2} &  71.8\tiny{$\pm$1.0} &  65.2\tiny{$\pm$0.4} &  68.9\tiny{$\pm$0.4} &  54.2\tiny{$\pm$1.3} &  64.1\tiny{$\pm$0.5} &  73.5\tiny{$\pm$0.6} & \cellcolor{gg}{66.3} & \textcolor{ForestGreen}{+9.3} \\
\multirow{-7}{*}{1B} & \multirow{-4}{*}{HaLoRA} & \cellcolor{gg}{2e-2} &  68.6\tiny{$\pm$1.8} &  60.2\tiny{$\pm$1.1} &  66.0\tiny{$\pm$0.7} &  50.8\tiny{$\pm$1.9} &  61.9\tiny{$\pm$1.0} &  71.1\tiny{$\pm$1.7} & \cellcolor{gg}{63.1} & \textcolor{ForestGreen}{+22.7} \\
 \midrule
 &  & 0 & 88.0 & 82.2 & 76.8 & 76.4 & 77.2 & 83.8 & 80.7 & -  \\
 &  & 5e-3 & 87.5\tiny{$\pm$0.2} & 81.6\tiny{$\pm$0.6} & 76.7\tiny{$\pm$0.3} & 75.7\tiny{$\pm$0.4} & 76.2\tiny{$\pm$1.4} & 80.5\tiny{$\pm$5.9} & 79.7 & - \\
 &  & 1e-2 & 86.1\tiny{$\pm$0.4} & 80.6\tiny{$\pm$1.1} & 76.3\tiny{$\pm$0.3} & 73.9\tiny{$\pm$0.6} & 71.5\tiny{$\pm$4.5} & 79.0\tiny{$\pm$4.7} & 77.9 & - \\
 & \multirow{-3}{*}{LoRA} & 2e-2 & 72.3\tiny{$\pm$9.2} & 65.2\tiny{$\pm$6.4} & 69.0\tiny{$\pm$2.9} & 60.1\tiny{$\pm$6.4} & 57.5\tiny{$\pm$8.4} & 65.5\tiny{$\pm$5.5} & 64.9 & - \\
  \cmidrule{2-11}
 &  & \cellcolor{gg}{0} &  87.3 &  81.4 &  77.7 &  76.7 &  80.6 &  84.2 & \cellcolor{gg}{81.3} & \textcolor{ForestGreen}{+0.6} \\
 &  & \cellcolor{gg}{5e-3} &  87.0\tiny{$\pm$0.1} &  82.3\tiny{$\pm$0.3} &  77.6\tiny{$\pm$0.3} &  76.4\tiny{$\pm$0.4} &  80.1\tiny{$\pm$0.1} &  83.9\tiny{$\pm$0.2} & \cellcolor{gg}{81.2} & \textcolor{ForestGreen}{+1.5} \\
\multirow{-8}{*}{LLaMA-3.2} &  & \cellcolor{gg}{1e-2} &  86.6\tiny{$\pm$0.1} &  81.6\tiny{$\pm$0.9} &  77.2\tiny{$\pm$0.3} &  76.0\tiny{$\pm$0.6} &  79.3\tiny{$\pm$0.5} &  83.4\tiny{$\pm$0.3} & \cellcolor{gg}{80.7} & \textcolor{ForestGreen}{+2.8} \\
\multirow{-7}{*}{3B} & \multirow{-4}{*}{HaLoRA} & \cellcolor{gg}{2e-2} &  84.6\tiny{$\pm$0.3} &  78.8\tiny{$\pm$0.6} &  75.8\tiny{$\pm$0.9} &  72.8\tiny{$\pm$1.5} &  77.2\tiny{$\pm$0.9} &  81.0\tiny{$\pm$1.1} & \cellcolor{gg}{78.4} & \textcolor{ForestGreen}{+13.5} \\
\bottomrule
\end{tabular}
}
\label{table_main}
\end{table*}

%% file: figures_tables/Table_compute_cost.tex
% \begin{table}[!t]
% \centering
% \caption{Computing overhead and performance during inference for LoRA and HaLoRA when fine-tuning LLaMA-3.2 1B.}
% \resizebox{0.9\linewidth}{!}
% {
% \begin{tabular}{lc|ccc}
% \toprule
% \multicolumn{2}{l|}{\textbf{Metric}} & \textbf{LoRA} & \textbf{HaLoRA} & $\Delta$ \\
% \midrule
% % \multicolumn{4}{c}{\textit{Training Stage}} \\
% % \midrule
% \multicolumn{2}{l|}{GPU Memory} & 23G & 25G & 8.6\% \\
% \multicolumn{2}{l|}{Training Time} & 2.5h & 2.7h & 8.0\% \\
% \midrule
% \multicolumn{5}{c}{\textit{Performance during Inference}} \\ 
% \midrule
% \multirow{4}{*}{Avg. Score} & \textit{-} & 62.3 & 67.6 & 8.5\% \\
% & \textit{5e-3} & 60.9 & 67.2 & 10.3\% \\
% & \textit{1e-2} & 57.0 & 66.3 & 16.3\% \\
% & \textit{2e-2} & 40.4 & 63.1 & 56.1\% \\
% \bottomrule
% \end{tabular}
% }
% \label{tab-cost}
% \end{table}

% \begin{table}[!t]
\begin{wraptable}{r}{7cm}
\vspace{-1em}
\centering
\caption{Computing overhead and performance of LoRA and proposed HaLoRA when finetuning Qwen2.5 0.5B.}
\begin{tabular}{lc|ccc}
\toprule
\multicolumn{2}{l|}{\textbf{Metric}} & \textbf{LoRA} & \textbf{HaLoRA} & $\Delta$ \\
\midrule
% \multicolumn{4}{c}{\textit{Training Stage}} \\
% \midrule
\multicolumn{2}{l|}{GPU Memory (G)} & 18.8 & 19.7 & +0.9 \\
\multicolumn{2}{l|}{Training Time (h)} & 1.83 & 1.95 & +0.12 \\
\midrule
\multicolumn{5}{c}{\textit{Performance during Inference}} \\ 
\midrule
\multirow{4}{*}{Avg. Score} & \textit{0} & 58.8 & 61.8 & +3.0 \\
& \textit{5e-3} & 56.7 & 61.5 & +4.8 \\
& \textit{1e-2} & 44.9 & 58.8 & +13.9 \\
& \textit{2e-2} & 28.1 & 48.6 & +20.5 \\
\bottomrule
\end{tabular}
\label{tab-cost}
% \end{table}
\end{wraptable}

%% file: revision/hardware_sim.tex
In this paper, we propose deploying finetuned LLMs on a hybrid CIM architecture, i.e., pretrained weights on RRAM and LoRA branches on SRAM.
RRAM-only CIM achieves high energy efficiency but suffers from inherent noise characteristics and complex write-verify operations \cite{ne_cim}, while SRAM-only CIM provides noise-free computation but is limited by its volatile nature and low storage density \cite{nn_cim}.
Therefore, we conduct hardware simulation on the energy costs, circuit area, and average performance comparing four strategies, including 1) both on GPU (A100), 2) both on RRAM, 3) both on SRAM, and 4) pretrained weights on RRAM and LoRA branches on SRAM.

For digital SRAM-based CIM, common implementations require 10 transistors to support computation for a single bit weight\cite{tu_ISSCC,10T_SRAM}.
In contrast, RRAM-based CIM enables storage and computation of one or multiple bits within the area of a single transistor \cite{chang_science,yao_nature}. 
\re{To compare area utilization among RRAM-only LoRA, SRAM-only LoRA, and HaLoRA implementations, we conducted a streamlined evaluation by multiplying computation density by the corresponding parameter quantity, using common 10T SRAM CIM cells and 1T1R RRAM cells as representative examples. Table \ref{table_set} shows the setting of the hardware simulations for transformer layers in LLaMA-3.2 1B model and LLaMA-3.2 3B model. The matrix size of the backbone weights is substantially larger than that of the LoRA branch, so the computation is primarily concentrated in the backbone. For RRAM-based analog CIM, prior work has demonstrated the potential for high-precision programming \cite{Thousands_level}. Accordingly, we assume that each weight can be stored in a single cell. We further adopt a commonly used deployment configuration with 8-bit weights and 8-bit activations, using 8-bit ADCs for readout.
With these configurations, we assessed the energy consumption of both RRAM and SRAM macros. 
Specifically, the energy breakdown of the RRAM macro includes energy consumption for arrays, ADCs, and other peripheral circuits (PCs), while accounting for the supporting dataflow of static vector-matrix multiplications and other dynamic operations.
This analysis was performed with XPEsim\cite{xpesim} and reported data of taped-out chips\cite{yao_nature,tu_ISSCC}.
}
\input{figures_tables/Table_sim_settings}

\begin{figure}[t]
    \centering
    \includegraphics[width=0.85\linewidth]{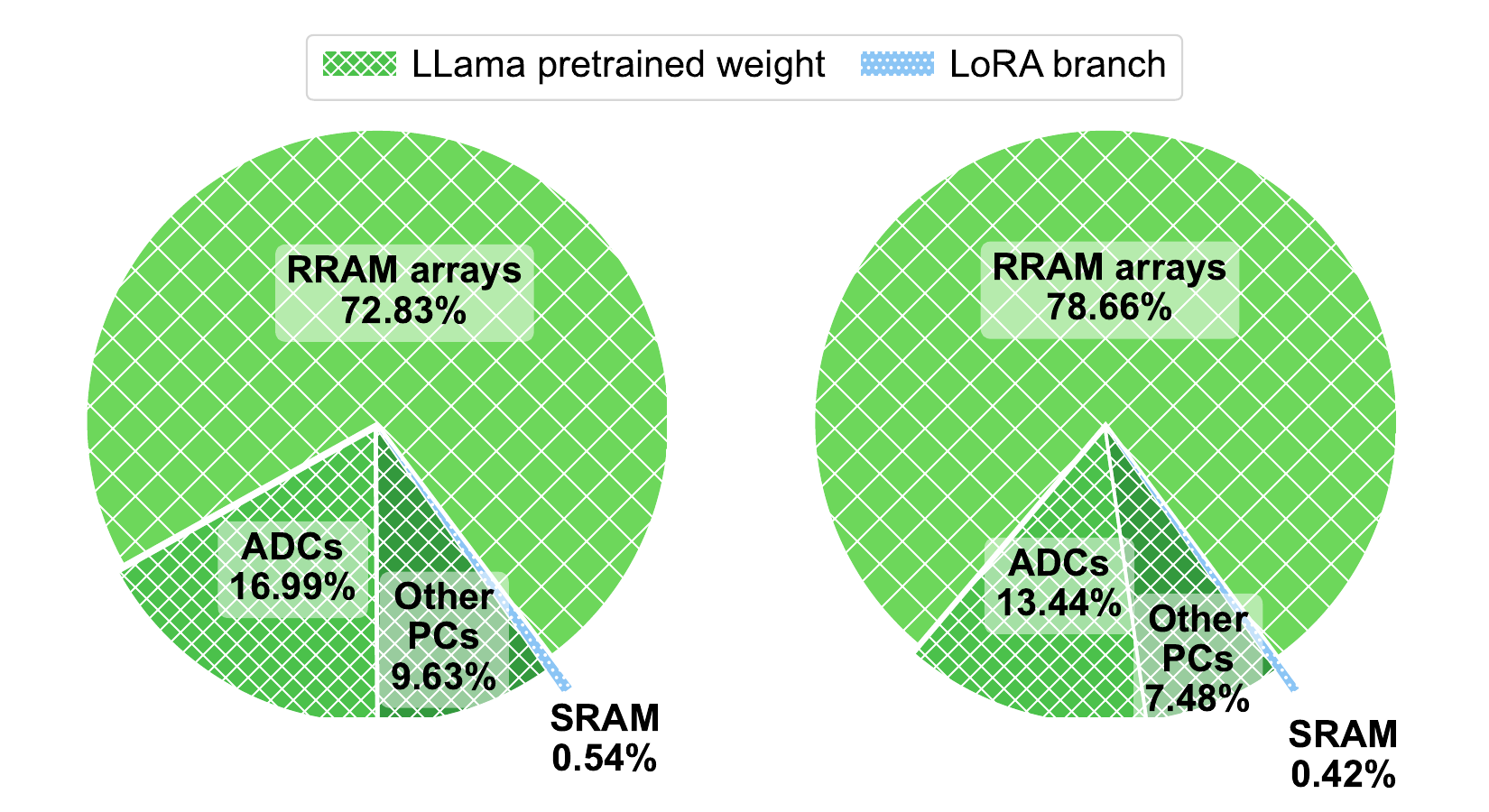}
    \caption{Energy breakdown of the transformer blocks in LLaMA-3.2 1B (left) and LLaMA-3.2 3B (right) for the hybrid CIM framework in HaLoRA.}
    \label{fig:breakdown}
\end{figure}

\input{figures_tables/Table_hardware}

We visualize the energy breakdown of the transformer blocks deployed on the hybrid CIM architecture shown in Fig. \ref{fig:breakdown} for a more detailed efficiency analysis.
RRAM arrays consume 72.83\% and 78.66\% of the energy for LLaMA-3.2 1B and 3B, respectively. This indicates that HaLoRA can efficiently leverage the energy-efficient characteristics of RRAM for LLaMA deployment. On the other hand, the energy consumption of LoRA branches implemented on SRAM accounts for only a small proportion (0.54\% and 0.42\%). This demonstrates that the additional overhead introduced by HaLoRA's LoRA branch deployment on SRAM is negligible compared to the RRAM-only LLaMA deployment, which further validates the efficiency of our design.

% As shown in Table \ref{tab-hard}, HaLoRA achieves a good trade-off between the hardware cost and performance for LLaMA-3.2 1B and 3B.
Table \ref{tab-hard} presents a comprehensive analysis demonstrating that HaLoRA achieves an excellent trade-off between hardware cost and performance for LLaMA-3.2 1B and 3B.
Specifically, our proposed HaLoRA framework requires 51.3 mJ energy cost for LLaMA-3.2 3B with 512 tokens, which is 3.0\% of the 1666.8 mJ from A100 and a slight 0.3\% more than 51.1 mJ from the RRAM-only strategy.
The circuit area of the hybrid HaLoRA framework is 10\% of the SRAM-only strategy and a slight 1.1\% more than RRAM-only.
Meanwhile, HaLoRA gets an average score of 81.2 with noise at level 5e-3, which is much higher than all three baseline strategies (i.e., 80.7 for LoRA (GPU/SRAM) and 79.7 for LoRA (RRAM)).
Regarding LLaMA-3.2 1B, the performance is consistent, demonstrating the effectiveness and robustness of HaLoRA.

%% file: figures_tables/Table_sim_settings.tex
\begin{table}[t]
\centering
\caption{\re{The hardware simulation settings of LLaMA-3.2 1B and LLaMA-3.2 3B.}}
\begin{tabular}{lcc}
\toprule
 & \textbf{LLaMA-3.2 1B} & \textbf{LLaMA-3.2 3B} \\
\midrule
Weight matrix sizes & 2048$\times$2048, 2048$\times$8192, & 3072$\times$3072, 3072$\times$8192, \\
 & 8192$\times$2048 & 8192$\times$3072 \\
 \midrule
LoRA branch sizes & 2048$\times$4, 4$\times$2048, 4$\times$8192, & 3072$\times$4, 4$\times$3072, 4$\times$8192, \\
 & 8192$\times$4, 4$\times$2048 & 8192$\times$4, 4$\times$3072 \\
 \midrule
Quantization setting & W8A8 & W8A8 \\
\midrule
Mapping approach & 1T1R-RRAM, Analog (8bit) & 1T1R-RRAM, Analog (8bit) \\
 & 10T-SRAM, Digital (1bit) & 10T-SRAM, Digital (1bit)\\
 \midrule
ADC ENOB & 8bit & 8bit \\
\bottomrule
\end{tabular}
\label{table_set}
\end{table}

%% file: figures_tables/Table_hardware.tex
% \begin{table}[!t]
% \centering
% \caption{
% Comparison among LoRA on A100 GPU, LoRA on RRAM, and proposed HaLoRA on hybrid CIM architecture when finetuning LLaMA-3.2 1B and 3B models.
% We report the energy cost of 512 tokens as input and benchmark performance without noise and with Gaussian noise at level 5e-3/1e-2/2e-2.
% }

% \resizebox{\linewidth}{!}
% {
% % \begin{tabular}{lc|ccc}
% % \toprule
% % \textbf{Method} & \textbf{Strategy} & \textbf{Efficiency} & \textbf{Performance} \\
% % \midrule
% % LoRA & GPU (A100) & 1.56 TOPS/W & 62.3 \\
% % LoRA & RRAM & 47.7 TOPS/W & 60.9/57.0/40.4 \\
% % \rowcolor{gg} HaLoRA & RRAM+SRAM & 47.4 TOPS/W & 67.2/66.3/63.1 \\
% % \bottomrule
% % \end{tabular}

% \renewcommand{\arraystretch}{1.5}
% \begin{tabular}{ll|cccc}
% \toprule
% \multirow{2}{*}{Model} & \textbf{Method} & \textbf{LoRA} & \textbf{LoRA} & \textbf{LoRA} & \textbf{HaLoRA (Ours)} \\
% \cmidrule{2-6}
% & \textit{Strategy} & \textit{GPU (A100)} & \textit{RRAM} & \textit{SRAM} & \textit{RRAM+SRAM} \\
% \midrule
% \multirow{-1}{*}{LLaMA} & \#Energy & 550.5 mJ  & 18.0 mJ & 36.0 mJ & 18.1 mJ \\
% \multirow{-1}{*}{3.2 1B} & \#Score & 62.3 & 60.9/57.0/40.4 & 62.3 & 67.2/66.3/63.1 \\
% & \#Area & - & 1x & 10x & 1.014x \\
% \midrule
% \multirow{-1}{*}{LLaMA} & \#Energe & 1666.8 mJ  & 51.1 mJ & 102.2 mJ & 51.3 mJ \\
% \multirow{-1}{*}{3.2 3B} & \#Score & 80.7 & 79.7/77.9/64.9 & 80.7 & 81.2/80.7/78.4 \\
% & \#Area & - & 1x & 10x & 1.011x \\
% \bottomrule
% \end{tabular}
% }
% \label{tab-hard}
% \end{table}

\begin{table}[!t]
\centering
% \begin{threeparttable}
\caption{
Comparison among LoRA on A100 GPU, LoRA on RRAM, LoRA on SRAM, and proposed HaLoRA on hybrid CIM architecture when finetuning LLaMA-3.2 1B and 3B models.
We report the energy cost of 512 tokens as input, circuit area ratio, and average performance without and with Gaussian noise at level 5e-3/1e-2/2e-2.
}
% \resizebox{\linewidth}{!}
% {
% \renewcommand{\arraystretch}{1.5}
\begin{tabular}{l|cccc}

\toprule
 \textbf{Method} & \textbf{LoRA} & \textbf{LoRA} & \textbf{LoRA} & \textbf{HaLoRA (Ours)} \\
\midrule
Strategy & GPU (A100) & RRAM & SRAM & RRAM+SRAM \\
\midrule
\multicolumn{5}{c}{\textit{LLaMA-3.2 1B}} \\
\midrule
 \#Energy $\downarrow$ & 550.5 mJ  & 18.0 mJ & 36.0 mJ & 18.1 mJ \\
  \#Area$^\dag$ $\downarrow$ & - & 1$\times$ & 10$\times$ & 1.014$\times$ \\
 \#Score $\uparrow$ & 62.3 & 60.9/57.0/40.4 & 62.3 & 67.2/66.3/63.1 \\

\midrule
\multicolumn{5}{c}{\textit{LLaMA-3.2 3B}} \\
\midrule
 \#Energy $\downarrow$ & 1666.8 mJ  & 51.1 mJ & 102.2 mJ & 51.3 mJ \\
  \#Area$^\dag$ $\downarrow$ & - & 1$\times$ & 10$\times$ & 1.011$\times$ \\
 \#Score $\uparrow$ & 80.7 & 79.7/77.9/64.9 & 80.7 & 81.2/80.7/78.4 \\

\bottomrule

\end{tabular}
% }
% \end{threeparttable}
\begin{tablenotes}
\footnotesize
\item  \ \ \ \ \ \ \ \ \ \ \ \ \ \ \ 
$\dag$Compared with the normalized transistor number needed for the CIM cell.
\end{tablenotes}
\label{tab-hard}
\end{table}

%% file: figures_tables/Table_case_study.tex
\begin{table}[!t]
\caption{Cases from LoRA-finetuned and HaLoRA-finetuned LLaMA-3.2 1B from the ARC-e datasets.
\textcolor{blue}{Blue} for correct answers and \textcolor{red}{red} for wrong.}

% \resizebox{\columnwidth}{!}
% {%
\begin{tabular}{lp{0.65\textwidth}}
\toprule
\multirow{2}{*}{\textbf{Input \#1}} & The Apollo 11 mission was able to retrieve samples of the Moon's surface because it was the first mission to have astronauts: \newline
Answer1: land on the Moon 
Answer2: orbit a planet 
Answer3: return to Earth
Answer4: walk in space \\
\midrule
Ground Truth & Answer1 \\
\midrule
\multicolumn{2}{c}{\textit{Noise-free}} \\
\midrule
LoRA & The correct answer is \textcolor{blue}{Answer1} \\
\textbf{HaLoRA (Ours)} & The correct answer is \textcolor{blue}{Answer1} \\
\midrule
\multicolumn{2}{c}{\textit{Noise ($\sigma$=2e-2)}} \\
\midrule
LoRA & \textcolor{red}{1/2/3/4} \\
\textbf{HaLoRA (Ours)} & The correct answer is \textcolor{blue}{Answer1} \\
\midrule
\midrule
\multirow{2}{*}{\textbf{Input \#2}} & Please choose the correct answer to the question: When trees develop leaves in the spring, 10 changes occur on the forest floor. Why does the development of leaves cause changes on the forest floor?
\newline
Answer1: Rainfall increases. 
Answer2: Sunlight is reduced.
Answer3: Wind speed increases.  
Answer4: Animal migration is stopped. \\
\midrule
Ground Truth & Answer2 \\
\midrule
\multicolumn{2}{c}{\textit{Noise-free}} \\
\midrule
LoRA & The correct answer is \textcolor{red}{Answer1} \\
\textbf{HaLoRA (Ours)} & The correct answer is \textcolor{blue}{Answer2} \\
\midrule
\multicolumn{2}{c}{\textit{Noise ($\sigma$=2e-2)}} \\
\midrule
LoRA & \textcolor{red}{Answer1/Answer2/Answer3/Answer4 is the correct answer. Answer1} \\
\textbf{HaLoRA (Ours)} & The correct answer is \textcolor{blue}{Answer2} \\
\bottomrule
\end{tabular}
% }
\label{tab_cases}
\end{table}

%% file: section/5-Conclusion.tex
\section{Conclusion}
In this work, we first present a hybrid CIM architecture for deploying LoRA-finetuned LLMs, utilizing both RRAM and SRAM architectures. 
Compared to RRAM-only and SRAM-only strategies, our hybrid CIM architecture achieves a good trade-off between hardware overhead and performance.
% Through theoretical analysis of device noise impact on LLM inference, we identify the performance degradation caused by RRAM non-ideality as a key challenge.
To address the performance degradation caused by RRAM non-ideality, we further propose HaLoRA, which minimizes the gap between optimization trajectories of the LoRA branch under both ideal and noisy conditions.
We theoretically show the upper bound of the gap and design an extra loss to enjoy both energy efficiency and accuracy during inference. 
% aligns the training objectives under both ideal and noisy conditions and optimizes the upper bound of the alignment loss. 
Experimental results demonstrate that HaLoRA consistently enhances LLM performance towards RRAM noises on six benchmarks when finetuning Qwen and LLaMA models, achieving up to 22.7 improvement in average score while maintaining
robustness at various noise types and noise levels.
For future work, we plan to extend HaLoRA to quantized LLMs, potentially combining HaLoRA with LoftQ \cite{li2023loftq}.
One possible way is to initialize the matrix in HaLoRA while being aware of the quantization gap.
Also, another interesting topic is to explore HaLoRA's effectiveness in more challenging tasks, such as mathematical reasoning \cite{amini2019mathqa} and code generation \cite{jiang2024survey}.